\documentclass{article}

\PassOptionsToPackage{numbers}{natbib}
\usepackage[preprint]{neurips_2021}
\usepackage{array}
    \newcolumntype{P}[1]{>{\centering\arraybackslash}p{#1}}
    \newcolumntype{M}[1]{>{\centering\arraybackslash}m{#1}}

\usepackage[utf8]{inputenc} 
\usepackage[T1]{fontenc}    
\usepackage{hyperref}       
\usepackage{url}            
\usepackage{booktabs}       
\usepackage{amsfonts}       
\usepackage{nicefrac}       
\usepackage{microtype}      
\usepackage{xcolor}         
\usepackage{amsmath,amssymb,amsthm}
\usepackage{url}
\usepackage{tikz}
\usepackage{subfigure}
\usepackage{multirow}
\usetikzlibrary{spy}
\usetikzlibrary{arrows,automata}
\usepackage{pgfplots}
\usepackage{mathrsfs}  
\usepackage{enumitem}
\usepackage{wrapfig}

\pgfplotsset{width=7cm,compat=1.8}

\usepackage{algorithm, algorithmic}

\newcommand{\X}{\mathbb{X}}
\newcommand{\Op}[1]{\operatorname{\mathcal{#1}}}

\newcommand{\Y}{\mathbb{Y}}
\newcommand{\R}{\mathcal{R}}
\newcommand{\G}{\mathcal{G}}
\newcommand{\W}{\mathbb{W}}

\newcommand{\Real}{\mathbb{R}}

\newcommand{\x}{\boldsymbol{x}}
\newcommand{\y}{\boldsymbol{y}}

\newcommand{\h}{\boldsymbol{h}}

\newtheorem{prop}{Proposition}

\newtheorem{thm}{Theorem}

\DeclareMathOperator*{\argmin}{arg\,min}
\DeclareMathOperator*{\argmax}{arg\,max}

\title{End-to-end reconstruction meets data-driven regularization for inverse problems}

\author{%
  Subhadip Mukherjee$^{*1}$, Marcello Carioni$^{*1}$, Ozan \"Oktem$^{2}$, and Carola-Bibiane Sch\"onlieb$^{1}$ \\
  $^1$Department of Applied Mathematics and Theoretical Physics, University of Cambridge, UK\\
  $^2$Department of Mathematics, KTH--Royal institute of Technology, Sweden \\
  $^*$Equal contribution authors\\
  Emails: \texttt{\{sm2467, mc2250, cbs31\}@cam.ac.uk, ozan@kth.se} \\
}

\begin{document}

\maketitle

\begin{abstract}
We propose an unsupervised approach for learning end-to-end reconstruction operators for ill-posed inverse problems. The proposed method combines the classical variational framework with iterative unrolling, which essentially seeks to minimize a weighted combination of the expected distortion in the measurement space and the Wasserstein-1 distance between the distributions of the reconstruction and ground-truth. More specifically, the regularizer in the variational setting is parametrized by a deep neural network and learned simultaneously with the unrolled reconstruction operator. The variational problem is then initialized with the reconstruction of the unrolled operator and solved iteratively till convergence. Notably, it takes significantly fewer iterations to converge, thanks to the excellent initialization obtained via the unrolled operator. The resulting approach combines the computational efficiency of end-to-end unrolled reconstruction with the well-posedness and noise-stability guarantees of the variational setting. Moreover, we demonstrate with the example of X-ray computed tomography (CT) that our approach outperforms state-of-the-art unsupervised methods, and that it outperforms or is on par with state-of-the-art supervised learned reconstruction approaches.

\end{abstract}

\section{Introduction}
Inverse problems are ubiquitous in imaging applications, wherein one seeks to recover an unknown model parameter $\x\in \X$ from its incomplete and noisy measurement, given by
\begin{align*}
\y^\delta =  \Op{A}(\x) + \boldsymbol{e}\in \Y.
\end{align*}
Here, the forward operator $\Op{A}: \X \rightarrow \Y$ models the measurement process in the absence of noise, and $\boldsymbol{e}$, with $\|\boldsymbol{e}\|_2\leq \delta$, denotes the measurement noise. For example, in computed tomography (CT), the forward operator computes line integrals of $\x$ over a predetermined set of lines in $\mathbb{R}^3$ and the goal is to reconstruct $\x$ from its projections along these lines. Without any further information about $\x$, inverse problems are typically ill-posed, meaning that there could be several reconstructions that are consistent with the measurement, even without any noise. 

The variational framework circumvents ill-posedness by encoding prior knowledge about $\x$ via a regularization functional $\R:\X\rightarrow \mathbb{R}$. In the variational setting, one solves  
\begin{equation}
\underset{\x \in \X}{\min}\,\, \mathcal{L}_{\Y}(\y^{\delta}, \Op{A}(\x)) + \lambda\, \R(\x),
\label{var_prob_def}
\end{equation}
where $\mathcal{L}_{\Y}:\Y\times \Y\rightarrow \mathbb{R}^+$ measures data-fidelity and $\R$ penalizes undesirable or unlikely solutions. The penalty $\lambda>0$  balances the regularization strength with the fidelity of the reconstruction. The variational problem \eqref{var_prob_def} is said to be well-posed if it has a unique solution varying continuously in $\y^{\delta}$.

The success of deep learning in recent years has led to a surge of data-driven approaches for solving inverse problems \cite{data_driven_inv_prob}, especially in imaging applications. 
These methods come broadly in two flavors: (i) end-to-end trained models that aim to directly map the measurement to the corresponding parameter and (ii) learned regularization methods that seek to find a data-adaptive regularizer instead of handcrafting it. Techniques in both categories have their relative advantages and demerits. Specifically, end-to-end approaches offer fast reconstruction of astounding quality, but lack in terms of theoretical guarantees and need supervised data (i.e., pairs of input and target images) for training. On the contrary, learned regularization methods inherit the provable well-posedness properties of the variational setting and can be trained in an unsupervised manner, however the reconstruction entails solving a high-dimensional optimization problem, which is often slow and computationally demanding. 

Our work derives ideas from learned optimization and adversarial machine learning, and makes an attempt to combine the best features of both aforementioned paradigms. In particular, the proposed method offers the flexibility of unsupervised training, produces fast reconstructions comparable to end-to-end supervised methods in quality, while enjoying the well-posedness and stability guarantees of the learned regularization framework. We first provide a brief overview of the literature on data-driven techniques for inverse problems before explaining our specific contributions in detail.    
\subsection{Related works}
End-to-end fully learned methods for imaging inverse problems either map the measurement directly to the image \cite{automap,Oh2018ETERnetET}, or learn to eliminate the artifacts from a model-based technique \cite{postprocessing_cnn}. Such approaches are data-intensive and may generalize poorly when trained on limited data. Iterative unrolling \cite{unrolling_eldar, admm_net, jonas_learned_iterative,kobler2017variational,meinhardt2017learning}, with its origin in the seminal work by Gregor and LeCun on data-driven sparse coding \cite{lecun_ista}, employs reconstruction networks that are inspired by optimization-based approaches and hence are interpretable. The unrolling paradigm enables one to encode the knowledge about the acquisition physics into the model architecture \cite{lpd_tmi}, thereby achieving data-efficiency. Nevertheless, end-to-end trained methods are supervised, and it is often challenging to obtain a large ensemble of paired data, especially in medical imaging applications.

Learned regularization methods, broadly speaking, aim to learn a data-driven regularizer in the variational setting. Some notable approaches in this paradigm include adversarial regularization (AR) \cite{ar_nips} and its convex counterpart \cite{acr_arxiv}, network Tikhonov (NETT) \cite{nett_paper}, total deep variation (TDV) \cite{kobler2020total}, etc., wherein one explicitly parametrizes the regularization functional using a neural network. The regularization by denoising (RED) approach aims to solve inverse problems by using a denoiser inside an algorithm for minimizing the variational objective \cite{red_schniter,romano2017RED,chan2016plug}. The Plug-and-play (PnP) method \cite{online_pnp_tci} with a learned denoiser is also implicitly equivalent to data-driven regularization, subject to additional constraints on the denoiser \cite{mmo_pesquet}. The deep image prior technique \cite{ulyanov2018deepImagePrior} does not require training, but it  seeks to regularize the solution by restricting it to be in the range of a deep generator and can thus be interpreted broadly as a deep learning-based regularization scheme. It is relatively easier to analyze learned regularization schemes using the machinery of classical functional analysis \cite{scherzer2009variational}, but they fall short in terms of reconstruction quality. Moreover, these methods require one to solve a high-dimensional, potentially non-convex, variational problem, leading to slow reconstruction and lack of provable convergence.    
\subsection{Specific contributions}
Our work seeks to combine iterative unrolling with data-adaptive regularization via an adversarial learning framework, and hence is referred to as unrolled adversarial regularization (UAR). The proposed approach learns a data-adaptive regularizer parametrized by a neural network, together with an iteratively unrolled reconstruction network that minimizes the corresponding expected variational loss in an adversarial setting. Unlike AR \cite{ar_nips} where the undesirable images are taken as the pseudo-inverse reconstruction and kept fixed throughout the training, we update them with the output of the unrolled reconstruction network in each training step, and, in turn, use them to further improve the regularizer. Thanks to the Kantorovich-Rubinstein (KR) duality \cite{wgan_main}, the alternating learning strategy of the reconstruction and the regularizer networks is equivalent to minimizing the expected data-fidelity over the distribution of the measurements, penalized by the Wasserstein-1 distance between the distribution of the reconstruction and the ground-truth. Once trained, the reconstruction operator produces a fast, end-to-end reconstruction. We show that this efficient reconstruction can be improved further by a refinement step that involves running a few iterations of gradient-descent on the variational loss with the corresponding regularizer, starting from this initial estimate. The refinement step not only produces reconstructions that outperform state-of-the-art unsupervised methods and are competitive with supervised methods, but also facilitates a well-posedness and stability analyses akin to classical variational approaches \cite{scherzer2009variational}. Our theoretical results on the learned unrolled operator and the regularizer are corroborated by strong experimental evidence for the CT inverse problem. 

\section{The proposed unrolled adversarial regularization (UAR) approach}
In this section, we give a short mathematical background on optimal transport, followed by a detailed description of the UAR framework, including the training protocol and the network architectures. 
\subsection{Background on Optimal transport}
Optimal transport theory \cite{cuturi2019, villani2009} has recently gained prominence in the context of measuring the distance between two probability distributions. In particular, given two probability distributions $\pi_1$ and $\pi_2$ on $\Real^n$, the Wasserstein-1 distance between them is defined as 
\begin{equation}
\W_1(\pi_1,\pi_2) := \inf_{\mu \in \Pi(\pi_1, \pi_2)} \int \|\x_1 - \x_2\|_2\, \mathrm{d}\mu(\x_1,\x_2),
\end{equation}
where $\Pi(\pi_1, \pi_2)$ denotes all transport plans having $\pi_1$ and $\pi_2$ as marginals. The Wasserstein distance has proven to be suitable for deep learning tasks, when the data is assumed to be concentrated on low-dimensional manifolds in $\Real^n$. It has been shown that in such cases, the Wasserstein distance provides a usable gradient during training \cite{wgan_main}, as opposed to other popular divergence measures.

By the KR duality, the Wasserstein-1 distance can be computed equivalently by solving a maximization problem over the space of 1-Lipschitz functions (denoted by $\mathbb{L}_1$) as 
\begin{equation}\label{eq:kantorovichduality}
\W_1(\pi_1,\pi_2) = \sup_{\R \in \mathbb{L}_1} \int \R(\x_1) \,\mathrm{d}\pi_1(\x_1) - \int \R(\x_2) \, \mathrm{d}\pi_2(\x_2),
\end{equation}
provided that $\pi_1$ and $\pi_2$ have compact support \cite{santambrogio2015}. 
Finally, we recall the definition of push-forward of probability measures, which is used extensively in our theoretical exposition. Given a probability measure $\pi$ on $\mathscr{A}$ and a measurable map $T: \mathscr{A} \rightarrow \mathscr{B}$, we define the push-forward of $\pi$ by $T$ (denoted as $T_\# \pi$) as a probability measure on $\mathscr{B}$ such that $T_\# \pi(B) = \pi(T^{-1}(B))$, for all measurable $B \subset \mathscr{B}$.
\subsection{Training strategy and model parametrization for UAR}
The principal idea behind UAR is to learn an unrolled deep  network $\G_{\phi}:\Y\rightarrow \X$ for reconstruction, together with a regularization functional $\R_{\theta}:\X \rightarrow{\mathbb{R}}$ parametrized by another convolutional neural network (CNN). The role of $\R_{\theta}$ is to discern ground-truth images from images produced by $\G_{\phi}$, while $\G_{\phi}$ learns to minimize the variational loss with $\R_{\theta}$ as the regularizer. As the images produced by $\G_{\phi}$ gets better, $\R_{\theta}$ faces a progressively harder task of telling them apart from the ground-truth images, thus leading to an improved regularizer. On the other hand, as the regularizer improves, the quality of reconstructions obtained using $\G_{\phi}$ improves simultaneously. Consequently, $\G_{\phi}$ and $\R_{\theta}$ helps each other improve as the training progresses via an alternating update scheme. 

\begin{algorithm}[t]
\caption{Learning unrolled adversarial regularization (UAR).}
\begin{algorithmic}
\STATE {\bf  1.} {\bf Input:} Data-set $\left\{\x_i\right\}_{i=1}^{N}\sim\pi_x$ and $\left\{\y_j\right\}_{j=1}^{N}\sim \pi_{y^\delta}$, initial reconstruction network parameter $\phi$ and regularizer parameter $\theta$, batch-size $n_{b}=1$, penalty $\lambda=0.1$, gradient penalty $\lambda_{\text{gp}}=10.0$, Adam optimizer parameters $(\beta_1,\beta_2)=(0.50,0.99)$. 
\STATE {\bf 2. (Learn a baseline regularizer)} {\bf for mini-batch $k=1,2,\cdots,10\,\frac{N}{n_b}$, do:}
    \begin{itemize}[leftmargin=0.2in]
        \item  Sample $\x_j\sim\pi_x$, $\y_j\sim \pi_{y^{\delta}}$, and $\epsilon_j\sim \text{uniform}\,[0,1]$; for $1 \leq j \leq n_b$. Compute $\boldsymbol{u}_j=\Op{A}^{\dagger}(\y_j)$ and $\x^{(\epsilon)}_j=\epsilon_j \x_j+\left(1-\epsilon_j\right)\boldsymbol{u}_j$.
        \item $\theta\leftarrow \text{Adam}_{\eta,\beta_1,\beta_2}(\theta,\nabla\,\tilde{J}_1(\theta))$, where $\eta=10^{-4}$, and $$\tilde{J}_1(\theta)=\frac{1}{n_b}\sum_{j=1}^{n_b}\left[\R_{\theta}\left(\x_j\right)-\R_{\theta}\left(\boldsymbol{u}_j\right)+\lambda_{\text{gp}}\left(\left\|\nabla \R_{\theta}\left(\x^{(\epsilon)}_j\right)\right\|_2-1\right)^2\right].$$
    \end{itemize}
\STATE {\bf 3. (Learn a baseline reconstruction operator)} {\bf for mini-batch $k=1,2,\cdots,5\,\frac{N}{n_b}$, do:}
    \begin{itemize}[leftmargin=0.2in]
        \item  Sample $\y_j\sim \pi_{y^\delta}$, and compute $\tilde{J}_2(\phi)=\frac{1}{n_b}\sum_{j=1}^{n_b}\left\|\y_j-\Op{A}\left(\G_{\phi}(\y_j)\right)\right\|_2^2+\lambda\,\R_{\theta}\left(\G_{\phi}(\y_j)\right)$.
        \item $\phi\leftarrow \text{Adam}_{\eta,\beta_1,\beta_2}(\phi,\nabla\,\tilde{J}_2(\phi))$, with $\eta=10^{-4}$.
    \end{itemize}
\STATE {\bf 4. (Jointly train $\R_{\theta}$ and $\G_{\phi}$ adversarially)} {\bf for mini-batch $k=1,2,\cdots,25\,\frac{N}{n_b}$, do:}
    \begin{itemize}[leftmargin=0.2in]
        \item  Sample $\x_j$, $\y_j$, and $\epsilon_j\sim \text{uniform}\,[0,1]$; for $1 \leq j \leq n_b$. Compute $\boldsymbol{u}_j=\G_{\phi}(\y_j)$ and $\x^{(\epsilon)}_j=\epsilon_j \x_j+\left(1-\epsilon_j\right)\boldsymbol{u}_j$.
        \item $\theta\leftarrow \text{Adam}_{\eta,\beta_1,\beta_2}(\theta,\nabla\,\tilde{J}_1(\theta))$, where $\tilde{J}_1(\theta)$ is as in Step 2, with $\eta=2\times 10^{-5}$.
        \item Update $\phi\leftarrow \text{Adam}_{\eta,\beta_1,\beta_2}(\phi,\nabla\,\tilde{J}_2(\phi))$ twice, with $\tilde{J}_2(\phi)$ as in Step 3, and $\eta=2\times 10^{-5}$.
    \end{itemize}
\STATE {\bf  5.} {\bf Output:} The trained networks $\G_{\phi}$ and $\R_{\theta}$.
\end{algorithmic}
\label{algo_uar_train}
\end{algorithm}
		
		
\subsubsection{Adversarial training} 
Let us denote by $\pi_x$ the ground-truth distribution and by $\pi_{y^\delta}$ the distribution of the noisy measurement. The UAR algorithm trains $\G_{\phi}$ and $\R_{\theta}$ simultaneously starting from an appropriate initialization. At the $k^{\text{th}}$ iteration of training, the parameters $\phi$ of the reconstruction network are updated as
\begin{equation}
  \phi_{k}\in \underset{\phi}{\argmin}\text{\,}J_k^{(1)}(\phi), \text{\,\,where\,\,}J_k^{(1)}(\phi):=\mathbb{E}_{\pi_{y^\delta}}\left[\left\|\Op{A}(\G_{\phi}(\y^{\delta}))-\y^{\delta} \right\|_2^2 + \lambda\text{\,}\R_{\theta_k}(\G_{\phi}(\y^{\delta}))\right],
    \label{phi_update}
\end{equation}
for a fixed regularizer parameter $\theta_k$. Subsequently, the regularizer parameters are updated as
\begin{eqnarray}
\theta_{k+1}\in\underset{\theta: \R_{\theta}\in \mathbb{L}_1}{\argmax}\,J_k^{(2)}(\theta), \text{\,\,where\,\,}J_k^{(2)}(\theta):=\mathbb{E}_{\pi_{y^\delta}}\left[\R_{\theta}\left(\G_{\phi_k}(\y^{\delta})\right)\right] - \mathbb{E}_{\pi_x}\left[\R_{\theta}\left(\x\right)\right].
\label{theta_update}
\end{eqnarray}
The learning protocol for UAR is unsupervised, since the loss functionals $J_k^{(1)}(\phi)$ and $J_k^{(2)}(\theta)$ can be computed based solely on the marginals $\pi_{x}$ and $\pi_{y^\delta}$. The alternating update algorithm in \eqref{phi_update} and \eqref{theta_update} essentially seeks to solve the min-max variational problem given by
\begin{align}\label{eq:adversarialloss}
\min_{\phi} \max_{\theta : \R_\theta \in \mathbb{L}_1} \mathbb{E}_{\pi_{y^\delta}}\left\|\Op{A}(\G_{\phi}(\y^{\delta}))-\y^{\delta} \right\|_2^2 + \lambda \left(\mathbb{E}_{\pi_{y^\delta}}\left[\R_{\theta}\left(\G_{\phi}(\y^{\delta})\right)\right] - \mathbb{E}_{\pi_x}\left[\R_{\theta}\left(\x\right)\right]\right).
\end{align}
Thanks to KR duality in \eqref{eq:kantorovichduality} and the definition of push-forward, \eqref{eq:adversarialloss} can be reformulated as 
\begin{align}\label{eq:wassersteinproblem}
\min_{\phi} \mathbb{E}_{\pi_{y^\delta}}\left\|\Op{A}(\G_{\phi}(\y^{\delta}))-\y^{\delta} \right\|_2^2 +  \lambda\W_1\left((\G_{\phi})_{\#}\pi_{y^\delta},\pi_x\right).
\end{align}
We refer the reader to Section \ref{sec:theoreticalanalysis} for a mathematically rigorous statement of this equivalence as well as for a well-posedness theory of the problem in \eqref{eq:wassersteinproblem}. 
Note that the equivalence of the alternating minimization procedure and the variational problem in \eqref{eq:wassersteinproblem} holds only if the regularizer is fully optimized in every iteration. Nevertheless, in practice, the reconstruction and regularizer networks are not fully optimized in every iteration. Instead, one refines the parameters by performing one (or a few) Adam updates on the corresponding loss functionals. Notably, if $\W_1\left((\G_{\phi_k})_{\#}\pi_{y^\delta},\pi_x\right)=0$, i.e., the parameters of $\G$ are such that the reconstructed images match the ground-truth in distribution, the loss functional $J_k^{(2)}(\theta)$ and its gradient vanish, leading to no further update of $\theta$. Thus, both networks stop updating when the outputs of $\G_{\phi}$ are indistinguishable from the ground-truth images. The concrete training steps are listed in Algorithm \ref{algo_uar_train}\footnote{Codes at \url{https://github.com/Subhadip-1/unrolling_meets_data_driven_regularization}.}.
\subsubsection{Iteratively unrolled reconstruction operator}
\label{g_phi_pdhg_sec}
The objective of $\G_{\phi}$ is to approximate the minimizer of the variational loss with $\R_{\theta}$ as the regularizer. Therefore, an iterative unrolling strategy akin to \cite{lpd_tmi} is adopted for parameterizing $\G_{\phi}$. Iterative unrolling seeks to mimic the variational minimizer via a primal-dual-style algorithm \cite{cham_pock}, with the proximal operators in the image and measurement spaces replaced with trainable CNNs. Although the variational loss in our case is non-convex, this parametrization for $\G_{\phi}$ is chosen because of its expressive power over a generic network. Initialized with $\x^{(0)}=\Op{A}^{\dagger}\y^{\delta}$ and $\h^{(0)}=\boldsymbol 0$, $\G_{\phi}$ produces a reconstruction $\x^{(L)}$ by iteratively applying the CNNs $\Lambda_{\phi_{\text{p}}^{(\ell)}}$ and $\Lambda_{\phi_{\text{d}}^{(\ell)}}$ in $\X$ and $\Y$, respectively:
\begin{equation*}
     \h^{(\ell+1)}=\Gamma_{\phi_{\text{d}}^{(\ell)}}\left(\h^{(\ell)},\sigma^{(\ell)}\Op{A}(\x^{\ell}),\y^{\delta}\right), \text{\,and\,}
    \x^{(\ell+1)}=\Lambda_{\phi_{\text{p}}^{(\ell)}}\left(\x^{(\ell)},\tau^{(\ell)}\Op{A}^*(\h^{\ell+1})\right), 0\leq\ell\leq L-1.
\end{equation*}
The step-size parameters $\sigma^{(\ell)}$ and $\tau^{(\ell)}$ are also made learnable and initialized as $\sigma^{(\ell)}=\tau^{(\ell)}=0.01$ for each layer $\ell$. The number of layers $L$ is typically much smaller (we take $L= 20$) than the number of iterations needed by an iterative primal-dual scheme to converge, thus expediting the reconstruction by two orders of magnitude once trained. 

The regularizer $\R_{\theta}$ is taken as a deep CNN with six convolutional layers, followed by one average-pooling and two dense layers in the end. 

\subsubsection{Variational regularization as a refinement step}
The unrolled operator $\G_{\phi^*}$ trained by solving the min-max problem in \eqref{eq:adversarialloss} provides reasonably good reconstruction when evaluated on X-ray CT, and already outperforms state-of-the-art unsupervised methods (c.f. Section \ref{numerics_sec}). We demonstrate that the regularizer $\R_{\theta^*}$ obtained together with $\G_{\phi^*}$ by solving \eqref{eq:adversarialloss} can be used in the variational framework to further improve the quality of the end-to-end reconstruction $\G_{\phi^*}(\y^{\delta})$ for a given $\y^{\delta}\in\Y$.
Specifically, we solve the variational problem 
\begin{align}\label{eq:inverseprob}
\min_{\x \in \mathbb{X}} \|\Op{A}(\x) - \y^{\delta}\|_2 + \lambda'\left( \R_{\theta^*}(\x) + \sigma \|\x\|^2_2\right),
\end{align} 
where $\lambda',\sigma\geq 0$, by applying gradient descent, initialized with $\G_{\phi^*}(\y^{\delta})$. The additional Tikhonov term in \eqref{eq:inverseprob} ensures coercivity of the overall regularizer, making it amenable to the standard well-posedness analysis \cite{scherzer2009variational}. Practically, it improves the stability of the gradient descent optimizer for \eqref{eq:inverseprob}. In practice, one essentially gets the same reconstruction with $\sigma=0$ subject to early stopping (100 iterations). Notably, the fidelity term in \eqref{eq:inverseprob} is the $\ell_2$ distance, instead of the squared-$\ell_2$ fidelity. We have empirically observed that this choice of the fidelity term improves the quality of the reconstruction, possibly due to the higher gradient of the objective in the initial solution $\G_{\phi^*}(\y^{\delta})$. Since the end-to-end reconstruction gives an excellent initial point, it takes significantly fewer iterations for gradiet-descent to recover the optimal solution to \eqref{eq:inverseprob}, and therefore UAR retains its edge in reconstruction time over fully variational approaches with learned regularizers (e.g., AR \cite{ar_nips} or its convex version \cite{acr_arxiv}).
\section{Theoretical results}\label{sec:theoreticalanalysis}
The theoretical properties of UAR are stated in this section and their proofs are provided in the supplementary document. Throughout this section, we assume that $\mathbb{X} = \Real^n$ and $\mathbb{Y} = \Real^k$, and
\begin{itemize}[leftmargin=0.3in]
\item[A1.] $\pi_x$ is compactly supported and $\pi_{y^\delta}$ is supported on a compact set $\mathcal{K}\subset \Real^k$ for every $\delta \geq 0$.
\end{itemize}
We then consider the following problem:
\begin{equation}
    \underset{\phi}{\inf} \,\underset{\R\in \mathbb{L}_1}{\sup}\,J_1\left(\G_{\phi},\R|\lambda,\pi_{y^\delta}\right):=\mathbb{E}_{\pi_{y^{\delta}}}\left\|\y^{\delta}-\Op{A}\G_\phi(\y^{\delta})\right\|_2^2+\lambda\,\left(\mathbb{E}_{\pi_{y^\delta}}\left[\R(\G_\phi(\y^{\delta}))\right]-\mathbb{E}_{\pi_x}\left[\R(\x)\right]\right).
    \label{eq:objectivetheoretical}
\end{equation}
Problem \eqref{eq:objectivetheoretical} is identical to the min-max variational problem defined in \eqref{eq:adversarialloss}, with the only difference that the maximization in $\R$ is performed over the space of all $1$-Lipschitz functions. Basically, we consider the theoretical limiting case where the neural networks $\R_\theta$ are expressive enough to approximate all functions in $\mathbb{L}_1$ with arbitrary accuracy. We make the following assumptions on $\G_\phi$:
\begin{itemize}[leftmargin=0.3in]
\item[A2.] $\G_\phi$ is parametrized over a finite dimensional compact set $K$, i.e. $\phi \in K$.
\item[A3.] $\G_{\phi_n} \rightarrow \G_\phi$ pointwise whenever $\phi_n \rightarrow \phi$.
\item[A4.] $\sup_{\phi \in K} \|\G_\phi\|_\infty < \infty$.
\end{itemize} 
Assumptions A2-A4 are satisfied, for instance, when $\G_\phi$ is parametrized by a neural network whose weights are kept bounded during training. These assumptions apply to all results in this section.
\subsection{Well-posedness of the adversarial loss}
Here, we prove well-posedness and stability to noise for the optimal reconstructions. As a consequence of the KR duality, \eqref{eq:objectivetheoretical} can be equivalently expressed as
\begin{equation}
    \underset{\phi}{\inf} \,J_2\left(\G_{\phi}|\lambda,\pi_{y^\delta}\right):=\mathbb{E}_{\pi_{y^\delta}}\left\|\y^{\delta}-\Op{A}\G_\phi(\y^{\delta})\right\|_2^2+\lambda \,\W_1(\pi_x,(\G_\phi)_{\#}\pi_{y^\delta})\,.
    \label{train_obj2}
\end{equation}
In the next theorem, we prove this equivalence, showing the existence of an optimal $\G_\phi$ and $\R$ for \eqref{eq:objectivetheoretical}.
\begin{thm}
\label{thm:well-posedness}
Problems \eqref{eq:objectivetheoretical} and \eqref{train_obj2} admit an optimal solution and
\begin{align}
\underset{\phi}{\inf} \,\,\underset{\R\in \mathbb{L}_1}{\sup}\,J_1\left(\G_{\phi},\R|\lambda,\pi_{y^\delta}\right)=\underset{\phi}{\inf} \,J_2\left(\G_{\phi}|\lambda,\pi_{y^\delta}\right). \label{eq:equivalencevalue}
\end{align}
Moreover, if $(\G_{\phi^*}, \R^*)$ is optimal for \eqref{eq:objectivetheoretical}, then $\G_{\phi^*}$ is optimal for \eqref{train_obj2}. Conversely, if $\G_{\phi^*}$ is optimal for \eqref{train_obj2}, then $(\G_{\phi^*}, \R^*)$ is optimal for \eqref{eq:objectivetheoretical}, for all $\R^* \in \argmax_{\R \in \mathbb{L}_1}\mathbb{E}_{\pi_{y^\delta}}\left[\R(\G_{\phi^*}(\y^{\delta}))\right]-\mathbb{E}_{\pi_x}\left[\R(\x)\right]$.
\end{thm}
Next, we study the stability of the optimal reconstruction $\G_{\phi^*}$ to noise. We consider $\G_{\phi_n}$, where
\begin{equation}
    \phi_n\in\underset{\phi}{\arg\,\inf}\,J_2\left(\G_{\phi}|\lambda,\pi_{y^{\delta_n}}\right), 
    \label{g_phi_n_def}
\end{equation} 
and show that $\G_{\phi_n}\rightarrow \G_{\phi^*}$ as $\delta_n\rightarrow\delta$, thus establishing noise-stability of the unrolled reconstruction. 
\begin{thm}[Stability to noise]\label{thm:stability}
Suppose, for given a sequence of noise levels $\delta_n \rightarrow \delta \in [0,\infty)$, it holds that $\pi_{y^{\delta_n}} \rightarrow \pi_{y^\delta}$ in total variation. Then, with $\phi_n$ as in \eqref{g_phi_n_def}, $\G_{\phi_n}\rightarrow \G_{\phi^*}$ up to sub-sequences. 
\end{thm}

\subsection{Effect of $\lambda$ on the end-to-end reconstruction}
\label{effect_of_lambda_sec}
In order to analyze the effect of the parameter $\lambda$ in \eqref{train_obj2} on the resulting reconstruction $\G_{\phi^*}$, it is convenient to introduce the following two sets:
\begin{align*}
\Phi_{\mathcal{L}}:=\left\{\phi: \mathbb{E}_{\pi_{y^{\delta}}}\left\|\y^{\delta}-\Op{A}\G_\phi(\y^{\delta})\right\|^2_2  = 0\right\} \text{\,\,and\,\,} \Phi_{\W}:=\left\{\phi: (\G_\phi)_\# \pi_{y^\delta} = \pi_x\right\}.
\end{align*}
We assume that both $\Phi_{\mathcal{L}}$ and $\Phi_{\W}$ are non-empty, which is tantamount to asking that the parametrization of the end-to-end reconstruction operator is expressive enough to approximate a right inverse of $\Op{A}$ ($\Phi_{\mathcal{L}} \neq \emptyset$) and a transport map from $\pi_{y^\delta}$ to $\pi_x$ ($\Phi_{\W} \neq \emptyset$), and therefore is not very restrictive (keeping in view the enormous approximation power of unrolled deep architectures).
\begin{prop}\label{thm:firstestimates}
Let $\G_{\phi^*}$ be a minimizer for \eqref{train_obj2}. Then, it holds that 
\begin{itemize}
    \item $\mathbb{E}_{\pi_{y}}\left\|\y^{\delta}-\Op{A}\G_{\phi^*}(\y^{\delta})\right\|^2_2 \leq \lambda \W_1(\pi_x,(\G_{\phi})_{\#}\pi_{y^\delta}),\quad \text{\,\,for every\,\,}\phi \in \Phi_{\mathcal{L}}$.
    \item $\displaystyle\W_1(\pi_x,(\G_{\phi^*})_{\#}\pi_{y^\delta}) \leq \frac{1}{\lambda} \mathbb{E}_{\pi_{y^\delta}}\left\|\y^{\delta}-\Op{A}\G_{\phi}(\y^{\delta})\right\|^2_2,\quad \text{\,\,for every\,\,} \phi \in \Phi_{\W}$.
\end{itemize}

\end{prop}
The previous proposition shows in a quantitative way that for small $\lambda$, the optimal $\G_{\phi^*}$ has less expected distortion in the measurement space as the quantity $\mathbb{E}_{\pi_{y^\delta}}\left\|\y^{\delta}-\Op{A}\G_{\phi^*}(\y^{\delta})\right\|^2_2$ is small. On the other hand, if $\lambda$ is large, then the optimal $\G_{\phi^*}$ maps $\pi_{y^{\delta}}$ is closer to $\pi_x$ as the quantity $\W(\pi_x,(\G_{\phi^*})_{\#}\pi_{y^\delta})$ is small. Therefore, the regularization is stronger in this case.

\begin{figure*}[t]
	\centering
	\subfigure[$\lambda=0.001$: 21.60, 0.21]{
		\includegraphics[height=1.3in]{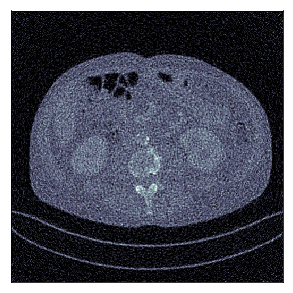}}
	\subfigure[$\lambda=0.01$: 25.33, 0.37]{
		\includegraphics[width=1.3in]{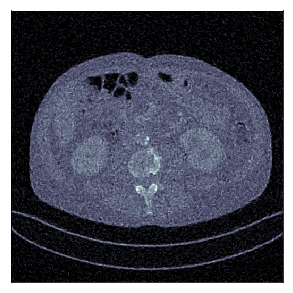}}
		\subfigure[$\lambda=0.1$: 34.65, 0.88]{
		\includegraphics[height=1.3in]{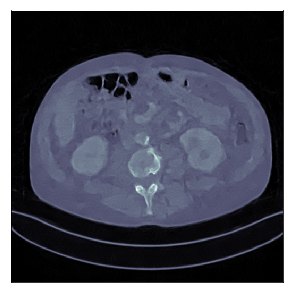}}
	\subfigure[$\lambda=1.0$: 33.96, 0.88]{
		\includegraphics[width=1.3in]{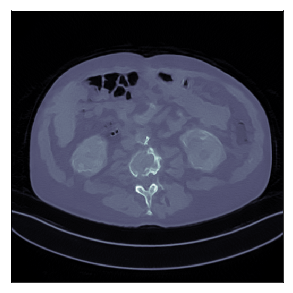}}
	\caption{Reconstruction of UAR for different $\lambda$. For $\lambda\rightarrow 0$, the unrolled generator seeks to find the minimizer of the expected data-fidelity loss, hence the reconstruction looks similar to FBP.}
	\label{effect_of_lambda_fig}
\end{figure*}

We extend this analysis by studying the behavior of the unrolled reconstruction as $\lambda$ converges to $0$ and to $+\infty$. Consider a sequence of parameters $\lambda_n > 0$ and the minimizer of the objective in \eqref{train_obj2} with parameter $\lambda_n$:
 \begin{equation}\label{eq:differentparameters}
\phi'_n\in\underset{\phi}{\arg \,\inf}\,\, J_2\left(\G_{\phi}|\lambda_n,\pi_{y^{\delta}}\right).
\end{equation}
\begin{thm}\label{thm:zero}
Let $\lambda_n \rightarrow 0$. Then, there exists $\phi_1^*\in \underset{\phi \in \Phi_{\mathcal{L}}}{\argmin}\,\, \W_1(\pi_x,(\G_\phi)_{\#}\pi_{y^\delta})$ such that $\G_{\phi'_n} \rightarrow \G_{\phi_1^*}$ up to sub-sequences, and $\underset{n \rightarrow \infty}{\lim}\,\,\frac{1}{\lambda_n}\underset{\phi}{\inf} \,\, J_2\left(\G_{\phi}|\lambda_n,\pi_{y^{\delta}}\right) =  \W_1(\pi_x,(\G_{\phi_1^*})_{\#}\pi_{y^\delta}).$
 
\end{thm}
\begin{thm}\label{thm:infinity}
Let $\lambda_n \rightarrow +\infty$. Then, there exists $\phi_2^* \in \underset{\phi \in \Phi_{\W}}{\argmin}\,\, \mathbb{E}_{\pi_{y^\delta}}\left\|\y^{\delta}-\Op{A}\G_\phi(\y^{\delta})\right\|^2_2$ such that $\G_{\phi'_n} \rightarrow \G_{\phi_2^*}$ up to sub-sequences, and $\underset{n \rightarrow \infty}{\lim}\,\,\underset{\phi}{\inf} \,\, J_2\left(\G_{\phi}|\lambda_n,\pi_{y^{\delta}}\right) =  \mathbb{E}_{\pi_{y^\delta}}\left\|\y^{\delta}-\Op{A}\G_{\phi_2^*}(\y^{\delta})\right\|^2_2$.
\end{thm}
Theorems \ref{thm:zero} and \ref{thm:infinity} characterize the optimal end-to-end reconstruction $\G_{\phi^*}$ as $\lambda \rightarrow 0$ and $\lambda \rightarrow \infty$, respectively. Specifically, if $\lambda \rightarrow 0$, $\G_{\phi^*}$ minimizes the Wasserstein distance between reconstruction and ground-truth among all the reconstruction operators that achieve zero expected data-distortion. In particular, $\G_{\phi^*}$ is close to the right inverse of $\Op{A}$ that minimizes the Wasserstein distance. Therefore, when $\lambda$ is very small, we expect to obtain a reconstruction that is close to the unregularized solution in quality. If $\lambda \rightarrow \infty$ on the other hand, the operator $\G_{\phi^*}$ is close to a transport map between $\pi_x$ and $\pi_{y^\delta}$, i.e., $(\G_{\phi^*})_\# \pi_{y^\delta} = \pi_x$, which minimizes the expected data-distortion. Therefore the reconstruction produces realistic images, but they are not consistent with the measurement. These theoretical observations are corroborated by the numerical results (c.f. Section \ref{numerics_sec}, Fig.~\ref{effect_of_lambda_fig}). One has to thus select a $\lambda$ that optimally trades-off data-distortion with the Wasserstein distance to achieve the best reconstruction performance. 
\begin{table}[t]
  \centering
  \begin{tabular}{l l c l r r l}
        \multicolumn{1}{l}{\textbf{method}} &
        & \multicolumn{1}{c}{\quad\textbf{PSNR (dB)}} 
        & \multicolumn{1}{c}{\quad\textbf{SSIM}} 
        & \multicolumn{1}{c}{\quad\textbf{\# param.}} 
        & \multicolumn{1}{c}{\quad\textbf{reconstruction}} \\
        &&&&& \textbf{time (ms)}\\
        \toprule
        FBP & & $21.28 \pm 0.13$  & $0.20 \pm 0.02$ & $1$ & $37.0 \pm 4.6$\\
        TV & & $30.31 \pm 0.52$  & $0.78 \pm 0.01$ & $1$ & $28371.4 \pm 1281.5$ \\
        \midrule
        \multicolumn{5}{l}{\emph{Supervised methods}} \\
        U-Net & & $34.50 \pm 0.65$  &  $0.90 \pm 0.01$ & $7215233$ & $44.4 \pm 12.5$ \\
        LPD & & $35.69 \pm 0.60$  & $0.91 \pm 0.01$  & $1138720$ & $279.8 \pm 12.8$ \\
        \midrule
        \multicolumn{5}{l}{\emph{Unsupervised methods}} \\
        AR & & $33.84 \pm 0.63$  & $0.86 \pm 0.01$ & $19338465$ & $22567.1 \pm 309.7$\\
        ACR  & & $31.55 \pm 0.54$  & $0.85 \pm 0.01$  &  $606610$ & $109952.4 \pm 497.8$\\
        \hline
        \multirow{3.5}{*}{UAR}& $\lambda=0.001$ & $21.59 \pm 0.11$ & $0.22 \pm 0.02$ & \multirow{3.5}{*}{20477186} & \multirow{3.5}{*} {$252.7 \pm 13.3$} \\
   & $\lambda=0.01$  & $25.25 \pm 0.08$ & $0.37 \pm 0.01$ & &  \\
  &  $\lambda=0.1$ & $34.35 \pm 0.66$  & $0.88 \pm 0.01$  &   &   \\
   & $\lambda=1.0$ & $33.27 \pm 0.76$ & $0.87 \pm 0.01$ & &  \\
   \hline
   UAR with   & $\lambda=\lambda'=0.1$ & $34.77 \pm 0.67$  & $0.90 \pm 0.01$ &  -- & $5863.3 \pm 106.1$ \\
   refinement &  &   &  &   &  \\
   \bottomrule
  \end{tabular}
  \\[2ex]
  \caption{Average PSNR and SSIM (with their standard deviations) for different reconstruction methods. The reconstruction times and the number of learnable parameters are also indicated. Without any refinement, UAR outperforms AR and ACR in reconstruction quality and reduces the reconstruction time by a couple of orders of magnitude. With the refinement, UAR becomes on par with supervised post-processing, but the reconstruction time is still four times smaller than AR.}
  \label{sparse_ct_table}
\end{table}
\subsection{End-to-end reconstruction vis-\`a-vis the variational solution}
The goal of this section is two-fold. Firstly, we theoretically justify the fact that the end-to-end reconstruction performs well, despite minimizing the expected loss over the distribution $\pi_{y^\delta}$. Secondly, we analyze the role of the regularizer in the variational setting in refining the end-to-end reconstruction. 

It is important to remark that the the end-to-end reconstruction is trained in on the expected variational loss computed using samples from $\pi_{y^\delta}$ and $\pi_x$. Therefore, the end-to-end reconstruction cannot learn a point-wise correspondence between measurement and model parameter, but only a distributional correspondence. Despite that, the end-to-end reconstruction achieves excellent performance for a given measurement vector $\y$. 
A justification of such phenomena is given by the next proposition. 
\begin{prop}\label{prop:pointwise}
Let $(\G_{\phi^*}, \R^*)$ be an optimal pair for \eqref{eq:objectivetheoretical} such that $\R^*\geq 0$ almost everywhere under $\left(\G_{\phi}\right)_{\#}\pi_{y^\delta}$. Define $M_1:=\mathbb{E}_{\pi_{y^\delta}}\left\|\y^{\delta}-\Op{A}\G_{\phi^*}(\y^{\delta})\right\|_2^2$ and $M_2:=\W_1(\pi_x,(\G_{\phi^*})_{\#}\pi_{y^\delta})$. Then, the following two upper bounds hold for every $\eta>0$:
\begin{itemize}
\item $\mathbb{P}_{\pi_{y^\delta}}\left\{ \y^{\delta} : \left\|\y^{\delta}-\Op{A}\G_{\phi^*}(\y^{\delta})\right\|^2_2  \geq \eta\right\}  \leq \frac{M_1}{\eta}.$
\item Suppose, $\R^*(\x) = 0$ for $\pi_x$-almost every $\x$. Then, $\mathbb{P}_{(\G_{
\phi^*})_\# \pi_{y^\delta}}\Big\{ \x : \R^*(\x) \geq \eta\Big\} \leq \frac{ M_2}{\eta}.$
\end{itemize}
\end{prop}
Proposition \ref{prop:pointwise} provides an estimate in probability of the sets $\{ \y^{\delta} : \left\|\y^{\delta}-\Op{A}\G_{\phi^*}(\y^{\delta})\right\|^2_2  \geq \eta\}$ and $ \{ \x : \R^*(\x) \geq \eta\}$. In particular, if $M_1$ is small, then $ \left\|\y^{\delta}-\Op{A}\G_{\phi^*}(\y^{\delta})\right\|^2_2$ is small in probability. If instead $M_2$ is small, then $\R^*(\x)$ is small in probability on the support of $(\G_{\phi^*})_\# \pi_{y^\delta}$, implying that samples $\G_{\phi^*}(\y^{\delta})$ are difficult to distinguish from the ground-truth.
We remark that the assumption $\R^*(\x) = 0$ can be justified using a data manifold assumption as in Section 3.3. of \cite{ar_nips}.
We now analyze the role of the regularizer $\R^*$ in the optimization of the variational problem \eqref{eq:inverseprob} that refines the end-to-end reconstruction $\G_{\phi^*}$.
We rely on a similar distributional analysis as the one performed in \cite{ar_nips}. For $\eta > 0$, consider the transformation by a gradient-descent step on $\R^*$ given by $g_\eta(\x) = \x - \eta \, \nabla \R^*(\x)$. Using the shorthand $\pi_{\G^*} := (\G_{\phi^*})_\# \pi_{y^\delta}$, and by denoting the distribution of $g_{\eta}(\x)$ as $\pi_{\eta}:=(g_\eta)_\# \pi_{\G^*}$ for $\x\sim \pi_{\G^*}$, we have the following theorem.
\begin{thm}[\cite{ar_nips}]\label{thm:wassgradientflow}
Suppose that $\eta \rightarrow \W_1(\pi_{\eta}, \pi_x )$ is differentiable at $\eta = 0$. Then, the derivative at $\eta=0$ satisfies $\frac{\mathrm{d}}{\mathrm{d}\eta} \W_1(\pi_{\eta},\pi_x ) \Big|_{\eta = 0}= - \mathbb{E}_{\pi_{\G^*}}\|\nabla \R^*(\x)\|_2^2$.
\end{thm}
This theorem states that a gradient-descent step performed on $\R^*$ at $\x=\G_{\phi^*}(\y^{\delta})$ decreases the Wasserstein distance with respect to the ground-truth distribution $\pi_x$. Therefore, if the gradient-descent step to solve the variational problem \eqref{eq:inverseprob} is initialized with the reconstruction $\G_{\phi^*}(\y^{\delta})$, the next iterate gets pushed closer to the ground-truth distribution $\pi_x$. We stress that this property holds because of the chosen initialization point, due to the relation between $\R^*$ and $\G_{\phi^*}$. For a different initialization, this property may not hold.
\begin{figure*}[t]
		\subfigure[ground-truth]{
		\includegraphics[width=1.3in]{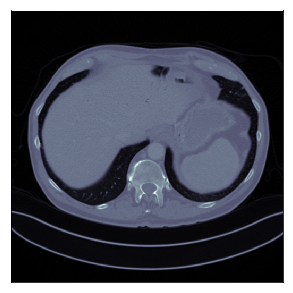}}
	\subfigure[FBP: 21.19 dB, 0.22]{
		\includegraphics[height=1.3in]{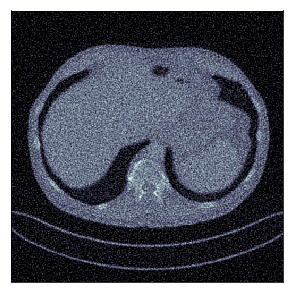}}
	\subfigure[TV: 29.85 dB, 0.79]{
		\includegraphics[width=1.3in]{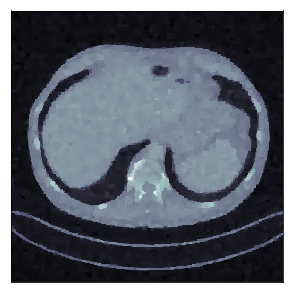}}
	\subfigure[U-net: 34.42 dB, 0.90]{
		\includegraphics[width=1.3in]{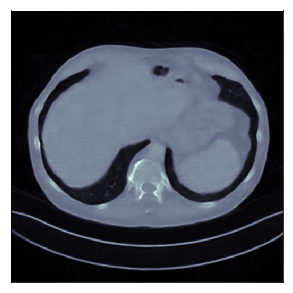}}\\
	\subfigure[LPD: 35.76 dB, 0.92]{
\includegraphics[width=1.3in]{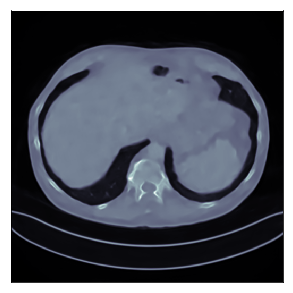}}
		\subfigure[ACR: 31.24 dB, 0.86]{
		\includegraphics[width=1.3in]{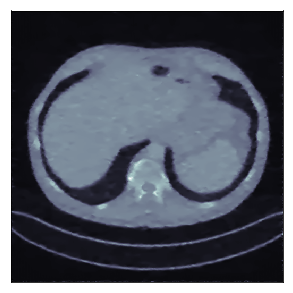}}
	\subfigure[AR: 33.52 dB, 0.86]{
		\includegraphics[width=1.3in]{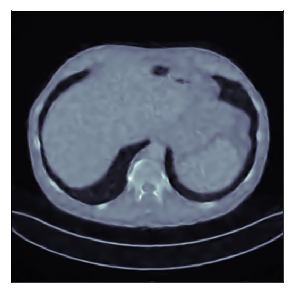}}
		\subfigure[UAR: 33.85 dB, 0.87]{
		\includegraphics[width=1.3in]{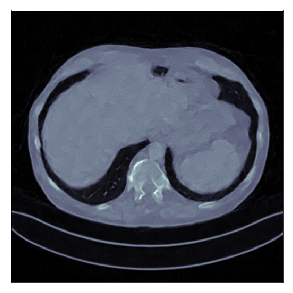}}
	\caption{Reconstruction on Mayo clinic data. UAR achieves better reconstruction quality than AR and ACR, while significantly reducing the reconstruction time (c.f. Table \ref{sparse_ct_table}). The reduction in reconstruction time comes at the expense of higher training complexity as compared to AR.}
	\label{ct_image_figure_mayo}
\end{figure*}
\begin{figure*}[t]
	\centering
	\subfigure[ground-truth]{
		\includegraphics[height=0.83in]{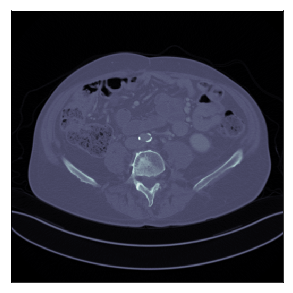}}
        \subfigure[34.94, 0.88]{
		\includegraphics[height=0.83in]{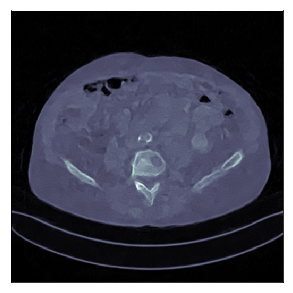}}
		\subfigure[35.46, 0.90]{
		\includegraphics[height=0.83in]{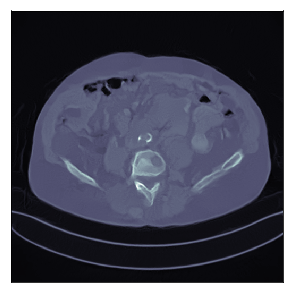}}
		\subfigure[ground-truth]{
		\includegraphics[height=0.83in]{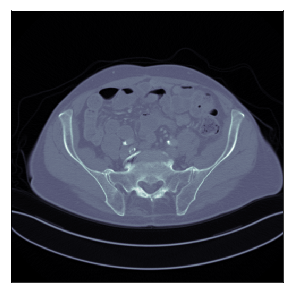}}
		\subfigure[33.84, 0.87]{
		\includegraphics[height=0.83in]{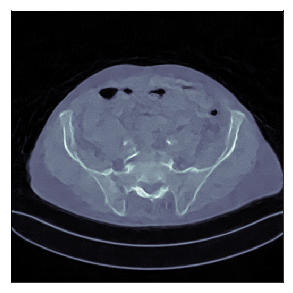}}
		\subfigure[34.24, 0.89]{
		\includegraphics[height=0.83in]{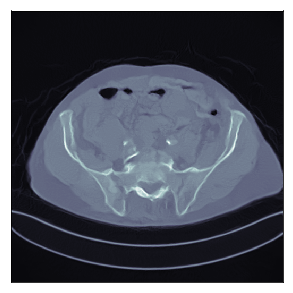}}
	\caption{Effect of refinement: (b) and (e): end-to-end reconstruction $\G_{\phi^*}(\y^{\delta})$; (c) and (f): the respective refined reconstructions. The PSNR (dB) and SSIM scores are indicated below.}
	\label{effect_of_refinement_fig}
\end{figure*}
\section{Numerical results}
\label{numerics_sec}
On the application front, we consider the prototypical inverse problem of CT reconstruction from noisy sparse-view projections. The abdominal CT scans for 10 patients, made publicly available by the Mayo-Clinic for the low-dose CT grand challenge \citep{mayo_ct_challenge}, were used in our numerical experiments. Specifically, 2250 2D slices of size $512 \times 512$ corresponding to 9 patients were used to train the models, while 128 slices from the remaining one patient were used for evaluation. The projections were simulated in ODL \cite{odl} using a parallel-beam geometry with 200 uniformly spaced angular positions of the source, with 400 lines per angle. Subsequently, Gaussian noise with standard deviation $\sigma_e=2.0$ was added to the projection data to simulate noisy sinograms.

The proposed UAR method is compared with two classical model-based approaches for CT, namely filtered back-projection (FBP) and total variation (TV). The LPD method \cite{lpd_tmi} and U-net-based post-processing \cite{postprocessing_cnn} of FBP are chosen as two supervised approaches for comparison. The AR approach \cite{ar_nips} and its convex variant \cite{acr_arxiv}, referred to as adversarial convex regularizer (ACR) , are taken as the competing unsupervised approaches. For LPD and AR, we develop a PyTorch-based implementation based on their publicly available TensorFlow codes\footnote{LPD: \url{https://github.com/adler-j/learned_primal_dual}.} \footnote{AR: \url{https://github.com/lunz-s/DeepAdverserialRegulariser}.}, while for ACR, we use the publicly available PyTorch implementation\footnote{ACR: \url{https://github.com/Subhadip-1/data_driven_convex_regularization}.}. 

The unrolled network $\G_{\phi}$ has 20 layers, with $5\times 5$ filters in both primal and dual spaces to increase the overall receptive field for sparse-view measurements. The hyper-parameters involved in training the UAR are specified in Algorithm \ref{algo_uar_train}. We found that first training a baseline regularizer and a corresponding baseline reconstruction operator helps stabilize the training process. Training the UAR model took approximately 30 hours on an NVIDIA Quadro RTX 6000 GPU (24 GB of memory).       

The average performance on the test images in terms of PSNR and SSIM \cite{ssim_paper_2004} indicates that UAR (with $\lambda=0.1$) outperforms AR and ACR by 0.3 dB and 2.6 dB, approximately. We would like to emphasize that this gain was found to be consistent across all test images and not just realized on average. With the refinement step, UAR surpasses AR by almost 0.7 dB and becomes on par with U-net post-processing. The end-to-end UAR reconstruction is a couple of orders of magnitude faster than AR, while the reduction in reconstruction time is by a factor of 4 with the refinement. The reconstructions of a representative test image using the competing methods are shown in Fig. \ref{ct_image_figure_mayo} for a visual comparison. The effect of $\lambda$ on the reconstruction of UAR is demonstrated in Fig. \ref{effect_of_lambda_fig}, which confirms the theoretical results in Section \ref{effect_of_lambda_sec}. The refinement step also visibly improves the reconstruction quality of the end-to-end operator, as shown in Fig. \ref{effect_of_refinement_fig}.

\section{Conclusions and limitations}
\label{conclude_sec}
To the best of our knowledge, this work makes the first attempt to blend end-to-end reconstruction with data-driven regularization via an adversarial learning framework. Our UAR approach retains the fast reconstruction of the former together with provable guarantees of the latter. We rigorously analyze the proposed framework in terms of well-posedness, noise-stability, and the effect of the regularization penalty, and establish a link between the trained reconstruction operator and the corresponding variational objective. We show strong numerical evidence of the efficacy of the UAR approach for CT reconstruction, wherein it achieves the same performance as supervised data-driven post-processing and outperforms competing unsupervised techniques. 
Our work paves the way to better understand the role of adversarially learned regularizers in solving ill-posed inverse problems, although several important aspects need further investigation. Since the learned regularizer is non-convex, the performance of gradient-descent on the variational objective greatly depends on initialization. This problem is partly addressed by the unrolled reconstruction operator that efficiently computes a better initial point for gradient descent. However, the precise relationship between the end-to-end reconstruction and the variational minimizer for a given measurement vector remains elusive. Moreover, the quality of the reconstruction relies on the expressive power of neural networks and thus suffers from the curse of dimensionality.
We believe that addressing such limitations will be important to better understand adversarial regularization methods.

\section{Acknowledgment}
MC acknowledges support from the Royal Society (Newton International Fellowship NIF\textbackslash R1\textbackslash 192048 Minimal partitions as a robustness boost for neural network classifiers).
CBS acknowledges support from the Philip Leverhulme Prize, the Royal Society Wolfson Fellowship, the EPSRC grants EP/S026045/1 and EP/T003553/1, EP/N014588/1, EP/T017961/1, the Wellcome Innovator Award RG98755, the Leverhulme Trust project Unveiling the invisible, the European Union Horizon 2020 research and innovation programme under the Marie Skodowska-Curie grant agreement No. 777826 NoMADS, the Cantab Capital Institute for the Mathematics of Information, and the Alan Turing Institute. SM acknowledges Thomas Buddenkotte for testing out the codes and the Wellcome Trust for funding and supporting his research.   

\bibliographystyle{plainnat}
\bibliography{bib}

\begin{thebibliography}{30}
\providecommand{\natexlab}[1]{#1}
\providecommand{\url}[1]{\texttt{#1}}
\expandafter\ifx\csname urlstyle\endcsname\relax
  \providecommand{\doi}[1]{doi: #1}\else
  \providecommand{\doi}{doi: \begingroup \urlstyle{rm}\Url}\fi

\bibitem[Adler et~al.(2017)Adler, Kohr, and \"Oktem]{odl}
J.~Adler, H.~Kohr, and O.~\"Oktem.
\newblock Operator discretization library (odl).
\newblock \emph{Software available from https://github.com/odlgroup/odl}, 2017.

\bibitem[Adler and \"Oktem(2009)]{jonas_learned_iterative}
Jonas Adler and Ozan \"Oktem.
\newblock Solving ill-posed inverse problems using iterative deep neural
  networks.
\newblock \emph{Inverse Problems}, 33\penalty0 (12), 2009.

\bibitem[Adler and {\"O}ktem(2018)]{lpd_tmi}
Jonas Adler and Ozan {\"O}ktem.
\newblock Learned primal-dual reconstruction.
\newblock \emph{IEEE transactions on medical imaging}, 37\penalty0
  (6):\penalty0 1322--1332, 2018.

\bibitem[Arjovsky et~al.(2017)Arjovsky, Chintala, and Bottou]{wgan_main}
Martin Arjovsky, Soumith Chintala, and L{\'e}on Bottou.
\newblock {W}asserstein generative adversarial networks.
\newblock In \emph{Proceedings of the 34th International Conference on Machine
  Learning}, pages 214--223, 2017.

\bibitem[Arridge et~al.(2019)Arridge, Maass, \"Oktem, and
  Sch\"onlieb]{data_driven_inv_prob}
Simon Arridge, Peter Maass, Ozan \"Oktem, and Carola-Bibiane Sch\"onlieb.
\newblock Solving inverse problems using data-driven models.
\newblock \emph{Acta Numerica}, 28:\penalty0 1--174, 2019.

\bibitem[Chambolle and Pock(2010)]{cham_pock}
A.~Chambolle and T.~Pock.
\newblock A first-order primal-dual algorithm for convex problems with
  applications to imaging.
\newblock \emph{J. Math. Imaging and Vision}, 40\penalty0 (1):\penalty0
  120--145, 2010.

\bibitem[Chan et~al.(2016)Chan, Wang, and Elgendy]{chan2016plug}
Stanley~H Chan, Xiran Wang, and Omar~A Elgendy.
\newblock Plug-and-play admm for image restoration: Fixed-point convergence and
  applications.
\newblock \emph{IEEE Transactions on Computational Imaging}, 3\penalty0
  (1):\penalty0 84--98, 2016.

\bibitem[Cuturi and Peyr\'{e}()]{cuturi2019}
M.~Cuturi and G.~Peyr\'{e}.
\newblock Computational {O}ptimal {T}ransport.
\newblock Arxiv preprint arXiv:1803.00567, 2019.
  \url{https://arxiv.org/pdf/1803.00567.pdf}.

\bibitem[Gregor and LeCun(2010)]{lecun_ista}
K.~Gregor and Y.~LeCun.
\newblock Learning fast approximations of sparse coding.
\newblock In \emph{Intl. Conf. on Machine Learning}, 2010.

\bibitem[{Jin} et~al.(2017){Jin}, {McCann}, {Froustey}, and
  {Unser}]{postprocessing_cnn}
K.~H. {Jin}, M.~T. {McCann}, E.~{Froustey}, and M.~{Unser}.
\newblock Deep convolutional neural network for inverse problems in imaging.
\newblock \emph{IEEE Transactions on Image Processing}, 26\penalty0
  (9):\penalty0 4509--4522, 2017.

\bibitem[Kobler et~al.(2017)Kobler, Klatzer, Hammernik, and
  Pock]{kobler2017variational}
Erich Kobler, Teresa Klatzer, Kerstin Hammernik, and Thomas Pock.
\newblock Variational networks: connecting variational methods and deep
  learning.
\newblock In \emph{German conference on pattern recognition}, pages 281--293.
  Springer, 2017.

\bibitem[Kobler et~al.(2020)Kobler, Effland, Kunisch, and
  Pock]{kobler2020total}
Erich Kobler, Alexander Effland, Karl Kunisch, and Thomas Pock.
\newblock Total deep variation for linear inverse problems.
\newblock In \emph{Proceedings of the IEEE Conference on Computer Vision and
  Pattern Recognition}, pages 7549--7558, 2020.

\bibitem[Li et~al.(2020)Li, Schwab, Antholzer, and Haltmeier]{nett_paper}
Housen Li, Johannes Schwab, Stephan Antholzer, and Markus Haltmeier.
\newblock {NETT}: solving inverse problems with deep neural networks.
\newblock \emph{Inverse Problems}, 36\penalty0 (6), 2020.

\bibitem[Lunz et~al.(2018)Lunz, {\"O}ktem, and Sch{\"o}nlieb]{ar_nips}
Sebastian Lunz, Ozan {\"O}ktem, and Carola-Bibiane Sch{\"o}nlieb.
\newblock Adversarial regularizers in inverse problems.
\newblock In \emph{Advances in Neural Information Processing Systems}, pages
  8507--8516, 2018.

\bibitem[McCollough(2014)]{mayo_ct_challenge}
C.~McCollough.
\newblock Tfg-207a-04: Overview of the low dose ct grand challenge.
\newblock \emph{Medical Physics}, 43\penalty0 (6):\penalty0 3759--3760, 2014.

\bibitem[Meinhardt et~al.(2017)Meinhardt, Moller, Hazirbas, and
  Cremers]{meinhardt2017learning}
Tim Meinhardt, Michael Moller, Caner Hazirbas, and Daniel Cremers.
\newblock Learning proximal operators: Using denoising networks for
  regularizing inverse imaging problems.
\newblock In \emph{Proceedings of the IEEE International Conference on Computer
  Vision}, pages 1781--1790, 2017.

\bibitem[Monga et~al.(2019)Monga, Li, and Eldar]{unrolling_eldar}
V.~Monga, Y.~Li, and Y.~Eldar.
\newblock Algorithm unrolling: Interpretable, efficient deep learning for
  signal and image processing.
\newblock \emph{arXiv preprint arXiv:1912.10557v3}, 2019.

\bibitem[Mukherjee et~al.(2021)Mukherjee, Dittmer, Shumaylov, Lunz, \"Oktem,
  and Sch\"onlieb]{acr_arxiv}
S.~Mukherjee, S.~Dittmer, Z.~Shumaylov, S.~Lunz, O.~\"Oktem, and C.-B.
  Sch\"onlieb.
\newblock Learned convex regularizers for inverse problems.
\newblock \emph{arXiv preprint arXiv:2008.02839v2}, 2021.

\bibitem[Oh et~al.(2018)Oh, Kim, Chung, Han, and Park]{Oh2018ETERnetET}
Changheun Oh, Dongchan Kim, Jun-Young Chung, Yeji Han, and H.~Park.
\newblock Eter-net: End to end mr image reconstruction using recurrent neural
  network.
\newblock In \emph{MLMIR@MICCAI}, 2018.

\bibitem[Pesquet et~al.(Apr. 2021)Pesquet, Repetti, Terris, and
  Wiaux]{mmo_pesquet}
J.-C. Pesquet, A.~Repetti, M.~Terris, and Y.~Wiaux.
\newblock Learning maximally monotone operators for image recovery.
\newblock \emph{arXiv preprint arXiv:2012.13247v2}, Apr. 2021.

\bibitem[Reehorst and Schniter(2019)]{red_schniter}
E.~T. Reehorst and P.~Schniter.
\newblock Regularization by denoising: clarifications and new interpretations.
\newblock \emph{IEEE Transactions on Computational Imaging}, 5\penalty0
  (1):\penalty0 52--67, 2019.

\bibitem[Romano et~al.(2017)Romano, Elad, and Milanfar]{romano2017RED}
Yaniv Romano, Michael Elad, and Peyman Milanfar.
\newblock The little engine that could: Regularization by denoising (red).
\newblock \emph{SIAM Journal on Imaging Sciences}, 10\penalty0 (4):\penalty0
  1804--1844, 2017.

\bibitem[Santambrogio(2015)]{santambrogio2015}
F.~Santambrogio.
\newblock \emph{Optimal {T}ransport for {A}pplied {M}athematicians}.
\newblock Birkh{\"a}user {B}asel, 2015.

\bibitem[Scherzer et~al.(2009)Scherzer, Grasmair, Grossauer, Haltmeier, and
  Lenzen]{scherzer2009variational}
Otmar Scherzer, Markus Grasmair, Harald Grossauer, Markus Haltmeier, and Frank
  Lenzen.
\newblock \emph{Variational methods in imaging}.
\newblock Springer, 2009.

\bibitem[Sun et~al.(2019)Sun, Wohlberg, and S.]{online_pnp_tci}
Y.~Sun, B.~Wohlberg, and Kamilov~U. S.
\newblock An online plug-and-play algorithm for regularized image
  reconstruction.
\newblock \emph{IEEE Transactions on Computational Imaging}, 5\penalty0
  (3):\penalty0 395--408, 2019.

\bibitem[Ulyanov et~al.(2018)Ulyanov, Vedaldi, and
  Lempitsky]{ulyanov2018deepImagePrior}
Dmitry Ulyanov, Andrea Vedaldi, and Victor Lempitsky.
\newblock Deep image prior.
\newblock In \emph{Proceedings of the IEEE Conference on Computer Vision and
  Pattern Recognition}, pages 9446--9454, 2018.

\bibitem[Villani(2009)]{villani2009}
C.~Villani.
\newblock \emph{Optimal transport --- Old and new}, volume 338 of
  \emph{Grundlehren der mathematischen Wissenschaften}.
\newblock Springer Berlin Heidelberg, 2009.
\newblock \doi{10.1007/978-3-540-71050-9}.

\bibitem[Wang et~al.(2004)Wang, Bovik, Sheikh, and Simoncelli]{ssim_paper_2004}
Z.~Wang, A.~C. Bovik, H.~R. Sheikh, and E.~P. Simoncelli.
\newblock Image quality assessment: From error visibility to structural
  similarity.
\newblock \emph{IEEE Transactions on Image Processing}, 13\penalty0
  (4):\penalty0 600--612, 2004.

\bibitem[Yang et~al.(2016)Yang, Sun, Li, and Xu]{admm_net}
Y.~Yang, J.~Sun, H.~Li, and Z.~Xu.
\newblock Deep admm-net for compressive sensing mri.
\newblock In \emph{Advances in Neural Information Processing Systems}, 2016.

\bibitem[Zhu et~al.(2018)Zhu, Liu, Cauley, Rosen, and Rosen]{automap}
B.~Zhu, J.~Z. Liu, S.~F. Cauley, B.~R. Rosen, and M.~S. Rosen.
\newblock Image reconstruction by domain-transform manifold learning.
\newblock \emph{Nature}, 555:\penalty0 487--492, 2018.

\end{thebibliography}

\appendix



\section{Proofs of the theoretical results}
In this section, we prove the theoretical results stated in Section \ref{sec:theoreticalanalysis}.
First, we recall the setting and the main definitions. For the set of assumptions used in this section, we refer to Assumptions A1 -- A4 stated in Section \ref{sec:theoreticalanalysis}.
The objective of the adversarial optimization is defined as
\begin{equation}
    \underset{\phi}{\inf} \,\underset{\R\in \mathbb{L}_1}{\sup}\,J_1\left(\G_{\phi},\R|\lambda,\pi_{y^\delta}\right):=\mathbb{E}_{\pi_{y^{\delta}}}\left\|\y^{\delta}-\Op{A}\G_\phi(\y^{\delta})\right\|_2^2+\lambda\,\left(\mathbb{E}_{\pi_{y^\delta}}\left[\R(\G_\phi(\y^{\delta}))\right]-\mathbb{E}_{\pi_x}\left[\R(\x)\right]\right).
    \label{eq:objectivetheoretical_app}
\end{equation}
In Section \ref{sec:theoreticalanalysis}, we claimed that the problem \eqref{eq:objectivetheoretical_app} is well-posed and is equivalent to
\begin{equation}
    \underset{\phi}{\inf} \,J_2\left(\G_{\phi}|\lambda,\pi_{y^\delta}\right):=\mathbb{E}_{\pi_{y^\delta}}\left\|\y^{\delta}-\Op{A}\G_\phi(\y^{\delta})\right\|_2^2+\lambda \,\W_1(\pi_x,(\G_\phi)_{\#}\pi_{y^\delta})\,.
    \label{train_obj2_app}
\end{equation}
This shows the connection between the training objective and the Wasserstein-$1$ distance between the ground-truth distribution and the distribution of the reconstruction. Here, we prove the theorems stated in Section \ref{sec:theoreticalanalysis} regarding well-posedness (Theorem \ref{thm:well-posedness}), stability to noise (Theorem \ref{thm:stability}), and dependence on the parameter $\lambda$ (Proposition \ref{thm:firstestimates}, Theorem \ref{thm:zero}, and Theorem \ref{thm:infinity}) for \eqref{eq:objectivetheoretical_app} and \eqref{train_obj2_app}. Moreover, we further discuss the relation between  \eqref{train_obj2_app} and the variational problem used as a refinement and prove Proposition \ref{prop:pointwise}.

We recall the dominated convergence theorem below, which is used as one of the main tools in our proofs. For the sake of completeness, we also recall the definition of narrow convergence of measures.  

\noindent \textbf{Dominated convergence theorem}: Consider a sequence of measurable functions $\left\{f_n\right\}_{n\in\mathbb{N}}$ defined on a measure space $(\Omega, \mathcal{F},\mu)$ such that $f_n \rightarrow f$ pointwise for a measurable function $f$ defined on $(\Omega, \mathcal{F},\mu)$. Suppose that for any $x\in \Omega$, $|f_n(x)|\leq g(x)$, where $\displaystyle\int_{\Omega} \, |g| \,\mathrm{d}\mu<\infty$. Then, it holds that 
\begin{align*}
\underset{n\rightarrow \infty}{\lim}\int_{\Omega} \, \left|f_n-f \right| \,\mathrm{d}\mu=0
\end{align*}
and consequently, $\displaystyle\underset{n\rightarrow \infty}{\lim}\int_{\Omega} \, f_n \,\mathrm{d}\mu=\int_{\Omega} \, f \,\mathrm{d}\mu$.

\noindent \textbf{Narrow convergence of measures}: Consider a sequence of measures $\left\{\mu_n\right\}_{n\in\mathbb{N}}$ defined on a measurable space $(\Omega, \mathcal{F})$. Given a measure $\mu$ defined on $(\Omega, \mathcal{F})$ we say that $\mu_n$ narrowly converges to $\mu$ if 
\begin{align*}
    \lim_{n \rightarrow +\infty} \int \varphi\, \mathrm{d}\mu_n = \int \varphi\, \mathrm{d}\mu
\end{align*}
for every $\varphi \in C_b(\Omega)$, where we denote by $C_b(\Omega)$ the set of bounded continuous functions on $\Omega$.

\subsection{Well-posedness of the adversarial loss: Proofs of Theorem \ref{thm:well-posedness} and Theorem \ref{thm:stability}}
\subsubsection{Proof of Theorem \ref{thm:well-posedness}}
We start by proving the existence of an optimal solution for \eqref{train_obj2_app}.
Let $\G_{\phi_n}$ be a minimizing sequence for \eqref{train_obj2_app}, namely a sequence of reconstruction operators such that 
\begin{align}\label{eq:minsequence}
\lim_{n\rightarrow +\infty} J_2\left(\G_{\phi_n}|\lambda,\pi_{y^\delta}\right) =   \underset{\phi}{\inf} \,J_2\left(\G_{\phi}|\lambda,\pi_{y^\delta}\right).
\end{align}
 As $\phi_n \in K$ and $K$ is compact and finite dimensional (see Assumption A2), there exists $\phi^* \in K$ such that, up to sub-sequences, $\phi_n \rightarrow \phi^*$ and consequently $\G_{\phi_n} \rightarrow \G_{\phi^*}$ pointwise (see Assumption A3). We now show that $\G_{\phi^*}$ is a minimum for \eqref{train_obj2_app}. Thanks to the continuity of $\Op{A}$, we know that $\left\|\y^{\delta}-\Op{A}\G_{\phi_n}(\y^{\delta})\right\|_2^2 \rightarrow \left\|\y^{\delta}-\Op{A}\G_{\phi^*}(\y^{\delta})\right\|_2^2$ pointwise. Moreover, using Assumptions A1 and A4, the bound
 \begin{align*}
 \left\|\y^{\delta}-\Op{A}\G_{\phi_n}(\y^{\delta})\right\|_2^2 \leq  \sup_{\y^{\delta} \in \text{supp}(\pi_{y^\delta})} 2\|\y^{\delta}\|^2_2 + 2\|\Op{A}\|^2_{\text{op}} \left(\sup_{\phi \in K} \|\G_{\phi}\|_\infty\right)^2 < \infty
 \end{align*}
holds for every $\y^{\delta} \in \text{supp}(\pi_{y^\delta})$, where we denote by $\|\Op{A}\|_{\text{op}}$ the operator norm of $\Op{A}$. Therefore, by applying the dominated convergence theorem, we obtain that
\begin{align}\label{eq:convfidelity}
\lim_{n \rightarrow +\infty} \mathbb{E}_{\pi_{y^{\delta}}}\left\|\y^{\delta}-\Op{A}\G_{\phi_n}(\y^{\delta})\right\|_2^2  = \mathbb{E}_{\pi_{y^{\delta}}}\left\|\y^{\delta}-\Op{A}\G_{\phi^*}(\y^{\delta})\right\|_2^2\,.
\end{align}
Notice now that for every $\varphi \in C_b(\Real^k)$.
\begin{align*}
\left|\int \varphi(\x)\, \mathrm{d} [(\G_{\phi_n})_\# \pi_{y^\delta}] - \varphi(\x)\, \mathrm{d}[(\G_{\phi^*})_\# \pi_{y^\delta}] \right| \leq \int |\varphi(\G_{\phi_n}(\y^{\delta}))-  \varphi(\G_{\phi^*}(\y^{\delta}))| \, \mathrm{d}\pi_{y^\delta} \rightarrow 0
\end{align*}
as $n\rightarrow +\infty$, using, again, the dominated convergence theorem together with Assumption A1. Thus, the probability measures $(\G_{\phi_n})_\# \pi_{y^\delta} $ converge narrowly to $(\G_{\phi^*})_\# \pi_{y^\delta}$ as $n\rightarrow +\infty$. Moreover, using again dominated convergence, together with the bound $\sup_{\phi \in K} \|\G_\phi\|_\infty < \infty$ (see Assumption A4), we also have
\begin{align}\label{eq:tightness}
\left|\int \|\x\|_2\, \mathrm{d} (\G_{\phi_n})_\# \pi_{y^\delta} - \|\x\|_2\, \mathrm{d} (\G_{\phi^*})_\# \pi_{y^\delta} \right| \leq \int |\|\G_{\phi_n}(\y^{\delta})\|_2-  \|\G_{\phi^*}(\y^{\delta})\|_2| \, \mathrm{d}\pi_{y^\delta} \rightarrow 0
\end{align}
as $n\rightarrow +\infty$. Thus, using \citep[Theorem 5.11]{santambrogio2015}, we infer that $\lim_{n \rightarrow +\infty} \W_1(\pi_x,(\G_{\phi_n})_{\#}\pi_{y^{\delta}}) = \W_1(\pi_x,(\G_{\phi^*})_{\#}\pi_{y^{\delta}})$. Finally using such convergence, together with \eqref{eq:convfidelity} and \eqref{eq:minsequence}, we conclude that
\begin{align*}
J_2\left(\G_{\phi^*}|\lambda,\pi_{y^\delta}\right) & =\mathbb{E}_{\pi_{y^{\delta}}}\left\|\y^{\delta}-\Op{A}\G_{\phi^*}(\y^{\delta})\right\|_2^2+\lambda \,\W_1(\pi_x,(\G_{\phi^*})_{\#}\pi_{y^{\delta}})\\
& = \lim_{n \rightarrow +\infty} \mathbb{E}_{\pi_{y^{\delta}}}\left\|\y^{\delta}-\Op{A}\G_{\phi_n}(\y^{\delta})\right\|_2^2+\lambda \,\W_1(\pi_x,(\G_{\phi_n})_{\#}\pi_{y^{\delta}})\\
& = \lim_{n \rightarrow +\infty} J_2\left(\G_{\phi_n}|\lambda,\pi_{y^\delta}\right) \\
& =\underset{\phi}{\inf} \,J_2\left(\G_{\phi}|\lambda,\pi_{y^\delta}\right),
\end{align*}
thus showing that $\G_{\phi^*}$ is a minimum for \eqref{train_obj2_app}.

We now show that \eqref{eq:objectivetheoretical_app} and \eqref{train_obj2_app} are equivalent.
Using the Kantorovich-Rubinstein duality \citep[Theorem 1.39]{santambrogio2015} and bound $\sup_{\phi \in K} \|\G_\phi\|_\infty < \infty$ (Assumption A4), we have that for every $\phi \in K$, there exists $\R^\phi \in \mathbb{L}_1$ such that 
\begin{align}
& \R^\phi \in \argmax_{\R \in \mathbb{L}_1} \left(\mathbb{E}_{\pi_{y^{\delta}}}\left[\R(\G_\phi(\y^{\delta}))\right]-\mathbb{E}_{\pi_x}\left[\R(\x)\right]\right),\,\text{and} \label{eq:KR1}\\ 
& \mathbb{E}_{\pi_{y^{\delta}}}\left[\R^\phi(\G_\phi(\y^{\delta}))\right]-\mathbb{E}_{\pi_x}\left[\R^\phi(\x)\right] = \W_1(\pi_x,(\G_\phi)_{\#}\pi_{y^{\delta}})\,.\label{eq:KR2}
\end{align}
Therefore, denoting by $\G_{\phi^*}$ the minimum for \eqref{train_obj2_app}, it holds that
\begin{align*}
   \underset{\phi}{\inf} & \,\underset{\R\in \mathbb{L}_1}{\sup}\,J_1\left(\G_{\phi},\R|\lambda,\pi_{y^\delta}\right)     \\
   & =  \underset{\phi}{\inf}  \,\mathbb{E}_{\pi_{y^{\delta}}}\left\|\y^{\delta}-\Op{A}\G_\phi(\y^{\delta})\right\|_2^2+\lambda \,\underset{\R\in \mathbb{L}_1}{\sup}\,\left(\mathbb{E}_{\pi_{y^{\delta}}}\left[\R(\G_\phi(\y^{\delta}))\right]-\mathbb{E}_{\pi_x}\left[\R(\x)\right]\right) \\
    & =    \underset{\phi}{\inf} \,\mathbb{E}_{\pi_{y^{\delta}}}\left\|\y^{\delta}-\Op{A}\G_\phi(\y^{\delta})\right\|_2^2+\lambda\,\left(\mathbb{E}_{\pi_{y^{\delta}}}\left[\R^\phi(\G_\phi(\y^{\delta}))\right]-\mathbb{E}_{\pi_x}\left[\R^\phi(\x)\right]\right) \\
   & =  \underset{\phi}{\inf} \,\mathbb{E}_{\pi_{y^{\delta}}}\left\|\y^{\delta}-\Op{A}\G_\phi(\y^{\delta})\right\|_2^2+\lambda\,\W_1(\pi_x,(\G_\phi)_{\#}\pi_{y^{\delta}})\\
  & = \mathbb{E}_{\pi_{y^{\delta}}}\left\|\y^{\delta}-\Op{A}\G_{\phi^*}(\y^{\delta})\right\|_2^2+\lambda\,\left(\mathbb{E}_{\pi_{y^{\delta}}}\left[\R^*(\G_{\phi^*}(\y^{\delta}))\right]-\mathbb{E}_{\pi_x}\left[\R^*(\x)\right]\right),
\end{align*}
where $\R^*$ is any $1$-Lipschitz function such that
\begin{align*}
& \R^* \in \argmax_{\R \in \mathbb{L}_1} \left(\mathbb{E}_{\pi_{y^{\delta}}}\left[\R(\G_{\phi^*}(\y^{\delta}))\right]-\mathbb{E}_{\pi_x}\left[\R(\x)\right]\right).
\end{align*}

In particular, \eqref{eq:equivalencevalue} in Theorem \ref{thm:well-posedness} holds and the pair $(\G_{\phi^*}, \R^*)$ is optimal for \eqref{eq:objectivetheoretical_app}. Viceversa, if $(\G_{\phi^*}, \R^*)$ is optimal for \eqref{eq:objectivetheoretical_app}, then for every $\hat \phi \in K$ we have
\begin{align*}
J_2\left(\G_{\phi^*}|\lambda,\pi_{y^\delta}\right) & = \underset{\R\in \mathbb{L}_1}{\sup}\,J_1\left(\G_{\phi^*},\R|\lambda,\pi_{y^\delta}\right) = 
\underset{\phi}{\inf} \,\underset{\R\in \mathbb{L}_1}{\sup}\,J_1\left(\G_{\phi},\R|\lambda,\pi_{y^\delta}\right) \\
& \leq \underset{\R\in \mathbb{L}_1}{\sup}\,J_1\left(\G_{\hat \phi},\R|\lambda,\pi_{y^\delta}\right) = J_2\left(\G_{\hat \phi}|\lambda,\pi_{y^\delta}\right)
\end{align*}
where we used the optimality of $(\G_{\phi^*}, \R^*)$ together with \eqref{eq:KR1} and \eqref{eq:KR2}, showing that $\G_{\phi^*}$ is a minimizer for \eqref{train_obj2_app}.
\qed

\subsubsection{Proof of Theorem \ref{thm:stability}}
Let $\delta_n$ be a sequence converging to $\delta$ as $n\rightarrow +\infty$ and 
\begin{equation}
  \phi_n\in\underset{\phi}{\arg\,\inf}\,J_2\left(\G_{\phi}|\lambda,\pi_{y^{\delta_n}}\right).
    \label{g_phi_n_def_app}
\end{equation}
Recall that $\pi_{y^{\delta_n}}$ converges in total variation to $\pi_{y^{\delta}}$. We denote this convergence by 
\begin{align}\label{eq:conv_tv_app}
\lim_{n \rightarrow+\infty}\|\pi_{y^{\delta_n}} - \pi_{y^{\delta}}\|_{\mathcal{M}} = 0\,.
\end{align}
Using the fact that $\phi_n \in K$ and $K$ is compact and finite dimensional, we know that there exists $\phi^* \in K$ such that $\phi_n \rightarrow  \phi^*$, up to sub-sequences. In particular, by Assumption A3, $\G_{\phi_n} \rightarrow
\G_{\phi^*}$, up to sub-sequences. We need to prove that 
\begin{equation}\label{eq:thesis_noise_app}
\phi^* \in \argmin_{\phi} \, J_2\left(\G_{\phi}|\lambda,\pi_{y^{\delta}}\right).
\end{equation}
First, notice that as $\pi_{y^{\delta_n}} \rightarrow \pi_{y^\delta}$ in total variation, it holds that for every bounded, measurable function $f$, 
\begin{align}\label{eq:totalvariationtest}
\int f \, \mathrm{d}\pi_{y^{\delta_n}} \rightarrow  \int f \, \mathrm{d}\pi_{y^{\delta}}\,.
\end{align}
Therefore, using the fact that $\G_\phi$ is bounded for every $\phi \in K$ (Assumption A4), $\Op{A}$ is linear, and the supports of $\pi_{y^\delta_n}$ and $\pi_{y^\delta}$ are uniformly contained in a common compact set (Assumption A1), it holds for every $\G_\phi$ that
\begin{align}\label{eq:convfix}
\lim_{n \rightarrow +\infty} \int \|\y^{\delta}  - \Op{A}\G_\phi(\y^{\delta})\|_2^2\, \mathrm{d}\pi_{y^{\delta_n}} = \int \|\y^{\delta}  - \Op{A}\G_\phi(\y^{\delta})\|_2^2\, \mathrm{d}\pi_{y^{\delta}}\,.
\end{align}
Moreover, thanks to the dominated convergence theorem, together with the pointwise convergence $\G_{\phi_n}\rightarrow \G_{\phi^*}$ and the uniform bound $\sup_n \|\G_{\phi_n}\|_\infty < \infty$ (Assumption A4), we have
\begin{align}\label{eq:domeasy}
\lim_{n\rightarrow +\infty} \int \|\y^{\delta}  - \Op{A}\G_{\phi_n}(\y^{\delta})\|_2^2\, \mathrm{d} \pi_{y^{\delta}} = \int \|\y^{\delta}  - \Op{A}\G_{\phi^*}(\y^{\delta})\|_2^2\, \mathrm{d} \pi_{y^{\delta}}\,.
\end{align}
Therefore 
\begin{align}
& \limsup_{n \rightarrow +\infty}\left| \int \|\y^{\delta} - \Op{A}\G_{\phi_n}(\y^{\delta})\|_2^2\, \mathrm{d}\pi_{y^{\delta_n}} - \int \|\y^{\delta}  - \Op{A}\G_{\phi^*}(\y^{\delta})\|_2^2\, \mathrm{d}\pi_{y^{\delta}} \right|\nonumber\\
& = \limsup_{n \rightarrow +\infty} \left| \int \|\y^{\delta}  - \Op{A}\G_{\phi_n}(\y^{\delta})\|_2^2\, \mathrm{d}(\pi_{y^{\delta_n}} - \pi_{y^{\delta}} + \pi_{y^{\delta}}) - \int \|\y^{\delta}  - \Op{A}\G_{\phi^*}(\y^{\delta})\|_2^2\, \mathrm{d}\pi_{y^{\delta}} \right|\nonumber\\
& \leq  \limsup_{n \rightarrow +\infty}\left| \int \|y  - \Op{A}\G_{\phi_n}(\y^{\delta})\|_2^2\, \mathrm{d}\pi_{y^{\delta}} - \int \|y  - \Op{A}\G_{\phi^*}(\y^{\delta})\|_2^2\, \mathrm{d}\pi_{y^{\delta}} \right|\nonumber \\
& \quad + \left| \int \|\y^{\delta}  - \Op{A}\G_{\phi_n}(\y^{\delta})\|_2^2\, \mathrm{d}(\pi_{y^{\delta}} - \pi_{y^{\delta_n}})  \right| \nonumber\\
& =  \limsup_{n \rightarrow +\infty}\left| \int \|\y^{\delta} - \Op{A}\G_{\phi_n}(\y^{\delta})\|_2^2\, \mathrm{d}(\pi_{y^{\delta}} - \pi_{y^{\delta_n}})  \right|\label{eq:beforelast1}  \\
& \leq \limsup_{n \rightarrow +\infty} \int 2\|\y^{\delta}\|_2^2  +  2(\|\Op{A}\|_{\text{op}} \sup_n\|\G_{\phi_n}\|_\infty)^2\, \mathrm{d}\|\pi_{y^{\delta}} - \pi_{y^{\delta_n}}\| \label{eq:beforelast2}  \\
& \leq  \Big[\sup_{\y^{\delta} \in \mathcal{K}} 2\|\y^{\delta}\|^2_2 + 2\|\Op{A}\|^2_{\text{op}} \Big(\sup_{\phi \in K} \|\G_{\phi}\|_\infty\Big)^2 \Big] \limsup_{n \rightarrow +\infty} \|\pi_{y^{\delta}} - \pi_{y^{\delta_n}}\|_{\mathcal{M}} =  0  \label{eq:beforelast3}
\end{align}
where in \eqref{eq:beforelast1} we use \eqref{eq:domeasy} and 
in \eqref{eq:beforelast2}--\eqref{eq:beforelast3}  we use \eqref{eq:conv_tv_app} together with the fact that $\G_{\phi_n}$ are uniformly bounded (Assumption A4), $\Op{A}$ is linear and the supports of $\pi_{y^\delta_n}$ and $\pi_{y^\delta}$ are uniformly contained in a common compact set $\mathcal{K}$  (Assumption A1). 

Consider now a test function $\varphi \in C_b(\Real^k)$. Notice that 
\begin{align*}
& \limsup_{n \rightarrow +\infty} \left|\int \varphi(\G_{\phi_n}(\y^{\delta}))\, \mathrm{d}\pi_{y^{\delta_n}} - \int \varphi(\G_{\phi^*}(\y^{\delta})) \, \mathrm{d}\pi_{y^{\delta}} \right| \\
& \leq \limsup_{n \rightarrow +\infty} \left|\int \varphi(\G_{\phi_n}(\y^{\delta}))\, \mathrm{d}(\pi_{y^{\delta_n}} - \pi_{y^{\delta}}) \right| +  \left|\int \varphi(\G_{\phi_n}(\y^{\delta}))\,\mathrm{d}\pi_{y^\delta} - \int \varphi(\G_{\phi^*}(\y^{\delta})) \, \mathrm{d}\pi_{y^\delta} \right|\\
& \leq \limsup_{n \rightarrow +\infty} \int |\varphi(\G_{\phi_n}(\y^{\delta}))|\, \mathrm{d}|\pi_{y^{\delta_n}} - \pi_{y^{\delta}}| +  \left|\int \varphi(\G_{\phi_n}(y))\,\mathrm{d}\pi_{y^\delta} - \int \varphi(\G_{\phi^*}(\y^{\delta})) \, \mathrm{d}\pi_{y^\delta} \right|\\
& \leq \limsup_{n \rightarrow +\infty} \|\varphi\|_\infty \|\pi_{y^{\delta_n}} - \pi_{y^{\delta}}\|_{\mathcal{M}} +  \left|\int \varphi(\G_{\phi_n}(\y^{\delta}))\,\mathrm{d}\pi_{y^\delta} - \int \varphi(\G_{\phi^*}(\y^{\delta})) \, \mathrm{d}\pi_{y^\delta} \right|\\
& = 0,
\end{align*}
where we use again \eqref{eq:conv_tv_app} together with the pointwise convergence $\G_{\phi_n} \rightarrow \G_{\phi^*}$ and the compactness of  the support of $\pi_{y^\delta}$. Such estimate prove that $(\G_{\phi_n})_\# \pi_{y^{\delta_n}}$ converges narrowly to $(\G_{\phi^*})_\# \pi_{y^{\delta}}$. Moreover, adapting the previous to test function $\varphi(\x) = \|\x\|_2$ and using additionally that $\sup_{n} \|\G_{\phi_n}\|_\infty < \infty$ (see Assumption A4) we infer 
\begin{align*}
\lim_{n \rightarrow +\infty} \left|\int \|\x\|_2\, \mathrm{d} [(\G_{\phi_n})_\# \pi_{y^{\delta_n}}] - \|\x\|_2\, \mathrm{d}[(\G_{\phi^*})_\# \pi_{y^\delta}] \right| =  0
\end{align*}
which, thanks to \citep[Theorem 5.11]{santambrogio2015} and together with the narrow convergence $(\G_{\phi_n})_\# \pi_{y^{\delta_n}} \rightarrow (\G_{\phi^*})_\# \pi_{y^{\delta}}$ implies 
\begin{equation}\label{eq:convwass}
\lim_{n \rightarrow +\infty} \W_1(\pi_x,(\G_{\phi_n})_\#\pi_{y^{\delta_n}}) = \W_1(\pi_x,(\G_{\phi^*})_{\#}\pi_{y^{\delta}})
\end{equation}
and similarly
\begin{equation}\label{eq:convwass2}
\lim_{n \rightarrow +\infty} \W_1(\pi_x,(\G_{\phi})_\#\pi_{y^{\delta_n}}) = \W_1(\pi_x,(\G_{\phi})_{\#}\pi_{y^{\delta}}) \quad \text{for all } \phi \in K.
\end{equation}
We are finally in position to prove \eqref{eq:thesis_noise_app}. Let $\phi \in K$ a competitor for the variational problem in \eqref{eq:thesis_noise_app}. Then thanks to the optimality of $\phi_n$
\begin{align*}
 J_2\left(\G_{\phi_n}|\lambda,\pi_{y^{\delta_n}}\right) \leq  J_2\left(\G_{\phi}|\lambda,\pi_{y^{\delta_n}}\right)
\end{align*}
for every $n$. Passing to the limit in the previous inequality using \eqref{eq:convwass}, \eqref{eq:convwass2}, \eqref{eq:convfix} and \eqref{eq:beforelast3} we obtain 
\begin{equation}
J_2\left(\G_{\phi^*}|\lambda,\pi_{y^{\delta}}\right) \leq  J_2\left(\G_{\phi}|\lambda,\pi_{y^{\delta}}\right)
\end{equation}
as we wanted to prove.
\qed

\subsection{Effect of $\lambda$ on the end-to-end reconstruction. Proofs of Proposition \ref{thm:firstestimates}, Theorem \ref{thm:zero} and Theorem \ref{thm:infinity}}

Here we prove Proposition \ref{thm:firstestimates}, Theorem \ref{thm:zero} and Theorem \ref{thm:infinity}. We remind the reader the definition of the function spaces 
\begin{align*}
\Phi_{\mathcal{L}}:=\left\{\phi: \mathbb{E}_{\pi_{y^{\delta}}}\left\|\y^{\delta}-\Op{A}\G_\phi(\y^{\delta})\right\|^2_2  = 0\right\} \text{\,\,and\,\,} \Phi_{\W}:=\left\{\phi: (\G_\phi)_\# \pi_{y^\delta} = \pi_x\right\}
\end{align*}
that we assume to be non-empty.

\subsubsection{Proof of Proposition \ref{thm:firstestimates}}
Let $\G_{\phi^*}$ be a minimizer for \eqref{train_obj2_app}. Then for every $\phi \in \Phi_{\mathcal{L}}$ we easily estimate
\begin{align*}
\mathbb{E}_{\pi_{y^{\delta}}}& \left\|\y^{\delta}-\Op{A}\G_{\phi^*}(\y^{\delta})\right\|^2_2 \\
& \leq \mathbb{E}_{\pi_{y^{\delta}}}\left\|\y^{\delta}-\Op{A}\G_{\phi}(\y^{\delta})\right\|^2_2  + \lambda \,\W_1(\pi_x,(\G_\phi)_{\#}\pi_{y^{\delta}})  -  \lambda \,\W_1(\pi_x,(\G_{\phi^*})_{\#}\pi_{y^{\delta}}) \\
& \leq  \lambda \,\W_1(\pi_x,(\G_\phi)_{\#}\pi_{y^{\delta}}),
\end{align*}
leading to the first estimate in Proposition \ref{thm:firstestimates}. Moreover, for every $\phi \in \Phi_{\W}$ we obtain the second estimate in Proposition \ref{thm:firstestimates}, that is
\begin{align*}
\lambda & \W_1(\pi_x,(\G_{\phi^*})_{\#}\pi_{y^{\delta}})  \\
& \leq \mathbb{E}_{\pi_{y^{\delta}}}\left\|\y^{\delta}-\Op{A}\G_{\phi}(\y^{\delta})\right\|^2_2  + \lambda \,\W_1(\pi_x,(\G_\phi)_{\#}\pi_{y^{\delta}}) - \mathbb{E}_{\pi_{y^{\delta}}}\left\|\y^{\delta}-\Op{A}\G(\y^{\delta})\right\|^2_2 \\
&\leq  \mathbb{E}_{\pi_{y^{\delta}}}\left\|\y^{\delta}-\Op{A}\G_{\phi}(\y^{\delta})\right\|^2_2 ,
\end{align*}
where we used that for every $\phi \in  \Phi_{\W}$, $\W_1(\pi_x, (\G_{\phi})_{\#}\pi_{y^{\delta}}) = 0$.
\qed

\subsubsection{Proof of Theorem \ref{thm:zero}} 
We are assuming $\lambda_n \rightarrow 0$ and 
 \begin{equation}\label{eq:differentparameters_app}
\phi'_n\in\underset{\phi}{\arg \,\inf}\,\, J_2\left(\G_{\phi}|\lambda_n,\pi_{y^{\delta}}\right).
\end{equation}
First, using the fact that $\phi'_n \in K$ and $K$  is compact and finite dimensional we know that there exists $\phi_1^* \in K$ such that $\phi'_n \rightarrow  \phi_1^*$, up to sub-sequences. In particular, it also holds that $\G_{\phi'_n} \rightarrow \G_{\phi_1^*}$, up to sub-sequences, by Assumption A3.
It remains to prove that 
\begin{align}\label{eq:minlambazero}
\phi_1^*\in \underset{\phi \in \Phi_{\mathcal{L}}}{\argmin}\,\, \W_1(\pi_x,(\G_\phi)_{\#}\pi_{y^\delta}).
\end{align}
First notice that by Proposition \ref{thm:firstestimates} we can select $\phi \in \Phi_{\mathcal{L}}$ such that 
\begin{align*}
\mathbb{E}_{\pi_{y^{\delta}}}\left\|\y^{\delta}-\Op{A}\G_{\phi_n'}(\y^{\delta})\right\|^2_2 \leq \lambda_n \W(\pi_x,(\G_{\phi})_{\#}\pi_{y^{\delta}})
\end{align*}
for every $n$. So, taking the limit for $n\rightarrow +\infty$ and using that $\lambda_n \rightarrow 0$ together with \eqref{eq:convfidelity} (where again we used Assumptions A1 and A4, and the dominated convergence theorem) we obtain $\mathbb{E}_{\pi_{y^{\delta}}}\left\|\y^{\delta}-\Op{A} \G_{\phi_1^*}(\y^{\delta})\right\|^2_2  = 0$. Now, let $\phi \in \Phi_{\mathcal{L}}$. Using \eqref{eq:differentparameters_app} we have that for every $n$ 
\begin{align}
\lambda_n \W_1(\pi_x, (\G_{\phi_n'})_\# \pi_{y^\delta}) & \leq \lambda_n \W_1(\pi_x, (\G_{\phi_n'})_\# \pi_{y^\delta}) +   \mathbb{E}_{\pi_{y^{\delta}}}\left\|\y^{\delta}-\Op{A}\G_{\phi_n'}(\y^{\delta})\right\|^2_2 \nonumber  \\
& \leq \lambda_n \W_1(\pi_x, (\G_{\phi})_\# \pi_{y^\delta}). \label{eq:lambdatozero}
\end{align}
With similar arguments as in the proof of Theorem \ref{thm:well-posedness} we can prove that the probability measures $(\G_{\phi'_n})_\# \pi_{y^\delta} $ converge narrowly to $(\G_{\phi_1^*})_\# \pi_{y^\delta}$ as $n\rightarrow +\infty$. Additionally using the bound $\sup_{\phi \in K} \|\G_\phi\|_\infty < \infty$ (see Assumption A4) we can repeat the computation in \eqref{eq:tightness} to prove that $\lim_{n \rightarrow +\infty} \W_1(\pi_x,(\G_{\phi'_n})_{\#}\pi_{y^{\delta}}) = \W_1(\pi_x,(\G_{\phi_1^*})_{\#}\pi_{y^{\delta}})$ \citep[Theorem 5.11]{santambrogio2015}. So, passing to the limit in \eqref{eq:lambdatozero} we conclude that 
\begin{align*}
\W_1(\pi_x, (\G_{\phi_1^*})_\# \pi_{y^\delta}) & = \lim_{n \rightarrow +\infty} \W_1(\pi_x, (\G_{\phi_n'})_\# \pi_{y^\delta}) \leq \W_1(\pi_x, (\G_{\phi})_\# \pi_{y^\delta}) 
\end{align*}
showing \eqref{eq:minlambazero}. We now prove the convergence $\underset{n \rightarrow \infty}{\lim}\,\,\frac{1}{\lambda_n}\underset{\phi}{\inf} \,\, J_2\left(\G_{\phi}|\lambda_n,\pi_{y^{\delta}}\right) =  \W_1(\pi_x,(\G_{\phi_1^*})_{\#}\pi_{y^\delta})$.
Notice that using that, as  $\mathbb{E}_{\pi_{y^{\delta}}}\left\|\y^{\delta}-\Op{A} \G_{\phi_1^*}(\y^{\delta})\right\|^2_2  = 0$ and \eqref{eq:differentparameters_app} we have
\begin{align*}
\frac{1}{\lambda_n}J_2\left(\G_{\phi_n'}|\lambda_n,\pi_{y^{\delta}}\right)\leq \W_1(\pi_x, (\G_{\phi_1^*})_\# \pi_{y^\delta})
\end{align*}
and trivially
\begin{align*}
\frac{1}{\lambda_n}J_2\left(\G_{\phi_n'}|\lambda_n,\pi_{y^{\delta}}\right)\geq  \W_1(\pi_x, (\G_{\phi_n'})_\# \pi_{y^\delta}).
\end{align*}
So, passing to the limit in the previous estimates and using that $\lim_{n \rightarrow +\infty} \W_1(\pi_x,(\G_{\phi'_n})_{\#}\pi_{y^{\delta}}) = \W_1(\pi_x,(\G_{\phi_1^*})_{\#}\pi_{y^{\delta}})$ we prove the desired convergence.
\qed

\subsubsection{Proof of Theorem \ref{thm:infinity}}
We are assuming $\lambda_n \rightarrow +\infty$ and 
 \begin{equation}\label{eq:differentparameters_app2}
\phi'_n\in\underset{\phi}{\arg \,\inf}\,\, J_2\left(\G_{\phi}|\lambda_n,\pi_{y^{\delta}}\right).
\end{equation}
First, using the fact that $\phi'_n \in K$ and $K$  is compact and finite dimensional we know that there exists $\phi_2^* \in K$ such that $\phi'_n \rightarrow  \phi_2^*$, up to sub-sequences. In particular, it also holds that $\G_{\phi'_n} \rightarrow \G_{\phi_2^*}$, up to sub-sequences, by Assumption A3.
It remains to prove that 
\begin{align}\label{eq:minlambazero2}
\phi_2^*\in \underset{\phi \in \Phi_{\W}}{\argmin}\,\, \mathbb{E}_{\pi_{y^\delta}}\left\|\y^{\delta}-\Op{A}\G_\phi(\y^{\delta})\right\|^2_2.
\end{align}
First notice that by Proposition \ref{thm:firstestimates} we can select $\phi \in \Phi_{\W}$ such that 
\begin{align}\label{eq:w1_app}
\W_1(\pi_x,(\G_{\phi_n'})_{\#}\pi_{y^\delta}) \leq \frac{\mathbb{E}_{\pi_{y^\delta}}\left\|\y^{\delta}-\Op{A}\G_{\phi}(\y^{\delta})\right\|^2_2}{\lambda_n}
\end{align}
for every $n$.  With similar arguments as in the proof of Theorem \ref{thm:well-posedness}, using Assumption A1 and Assumption A4 together with \citep[Theorem 5.11]{santambrogio2015} there holds that $\lim_{n \rightarrow +\infty} \W_1(\pi_x,(\G_{\phi'_n})_{\#}\pi_{y^{\delta}}) = \W_1(\pi_x,(\G_{\phi_2^*})_{\#}\pi_{y^{\delta}})$.
So, taking the limit in \eqref{eq:w1_app} for $n\rightarrow +\infty$ and using that $\lambda_n \rightarrow +\infty$ we obtain $\W_1(\pi_x,(\G_{\phi_2^*})_{\#}\pi_{y^\delta}) = 0$.

Let now $\phi \in \Phi_{\W}$. Using \eqref{eq:differentparameters_app2} we have that for every $n$ 
\begin{align}
\mathbb{E}_{\pi_{y^{\delta}}}\left\|\y^{\delta}-\Op{A} \G_{\phi_n'}(\y^{\delta})\right\|^2_2  & \leq   \mathbb{E}_{\pi_{y^{\delta}}}\left\|\y^{\delta}-\Op{A}\G_{\phi_n'}(\y^{\delta})\right\|^2_2 + \lambda_n \W_1(\pi_x, (\G_{\phi_n'})_\# \pi_{y^\delta}) \nonumber  \\
& \leq  \mathbb{E}_{\pi_{y^{\delta}}}\left\|\y^{\delta}-\Op{A}\G_{\phi}(\y^{\delta})\right\|^2_2. \label{eq:lambdatoinf}
\end{align}
With similar arguments as in the proof of Theorem \ref{thm:well-posedness}, using dominated convergence theorem together with  Assumption A1 and Assumption A4 it holds that
\begin{align}\label{eq:convfidelity_app}
\lim_{n \rightarrow +\infty} \mathbb{E}_{\pi_{y^{\delta}}}\left\|\y^{\delta}-\Op{A}\G_{\phi'_n}(\y^{\delta})\right\|_2^2  = \mathbb{E}_{\pi_{y^{\delta}}}\left\|\y^{\delta}-\Op{A}\G_{\phi_2^*}(\y^{\delta})\right\|_2^2\,.
\end{align}
So, passing to the limit in \eqref{eq:lambdatoinf} we conclude that 
\begin{align*}
\mathbb{E}_{\pi_{y^{\delta}}}\left\|\y^{\delta}-\Op{A} \G_{\phi_2^*}(\y^{\delta})\right\|^2_2  & = \lim_{n \rightarrow +\infty} \mathbb{E}_{\pi_{y^{\delta}}}\left\|\y^{\delta}-\Op{A} \G_{\phi_n'}(\y^{\delta})\right\|^2_2 \leq \mathbb{E}_{\pi_{y^{\delta}}}\left\|\y^{\delta}-\Op{A} \G_{\phi}(\y^{\delta})\right\|^2_2 
\end{align*}
showing \eqref{eq:minlambazero2}. 

We now show the convergence $\underset{n \rightarrow \infty}{\lim}\,\,\underset{\phi}{\inf} \,\, J_2\left(\G_{\phi}|\lambda_n,\pi_{y^{\delta}}\right) =  \mathbb{E}_{\pi_{y^\delta}}\left\|\y^{\delta}-\Op{A}\G_{\phi_2^*}(\y^{\delta})\right\|^2_2$. Using that, $\W_1(\pi_x,(\G_{\phi_2^*})_{\#}\pi_{y^\delta}) = 0$, together with \eqref{eq:differentparameters_app2} we have
\begin{align*}
J_2\left(\G_{\phi_n'}|\lambda_n,\pi_{y^{\delta}}\right)\leq  \mathbb{E}_{\pi_{y^{\delta}}}\left\|\y^{\delta}-\Op{A} \G_{\phi_2^*}(\y^{\delta})\right\|^2_2
\end{align*}
and trivially
\begin{align*}
J_2\left(\G_{\phi_n'}|\lambda_n,\pi_{y^{\delta}}\right) \geq  \mathbb{E}_{\pi_{y^{\delta}}}\left\|\y^{\delta}-\Op{A} \G_{\phi_n'}(\y^{\delta})\right\|^2_2.
\end{align*}
So, passing to the limit in the previous estimate and using \eqref{eq:convfidelity_app} we prove the desired convergence.
\qed

\subsection{End-to-end reconstruction vis-\`a-vis the variational solution. Proof of Proposition \ref{prop:pointwise} and further discussion}

\subsubsection{Proof of Proposition \ref{prop:pointwise}}
The first upper bound in Proposition \ref{prop:pointwise} is a simple application of Markov inequality for probability measures, which states that every non-negative random variable $U$ satisfies $\mathbb{P}\left(U\geq \eta\right)\leq \frac{\mathbb{E}[U]}{\eta}$, for any $\eta>0$.

For the second upper bound notice that using Theorem \ref{thm:well-posedness} we have
\begin{align}
\int \R^*(\x)\, \mathrm{d}[(\G_{\phi^*})_\#\pi_{y^\delta}] = \int \R^*(\x)\, \mathrm{d}[(\G_{\phi^*})_\#\pi_{y^\delta}] - \int \R^*(\x)\, \mathrm{d}\pi_x =   \W_1(\pi_x,(\G_{\phi^*})_{\#}\pi_{y^\delta}) 
\end{align}
where we also use the assumption: $\R^*(\x) = 0$ for $\pi_x$-almost every $\x$. Therefore the second upper bound in Proposition \ref{prop:pointwise} follows from an application of Markov inequality, thanks to the assumed positivity of $\R^*$.
\qed

We remark the assumption regarding the positivity of $\R^*$ is not restrictive, as $\R^* + C$ is optimal for every $C \in \Real$. However,  it is not always true that $\R^*(\x) = 0$ for $\pi_x$-almost every $\x$. As discussed in Section \ref{sec:theoreticalanalysis}, such assumption can be justified using a suitable weak manifold assumption for $\pi_x$.

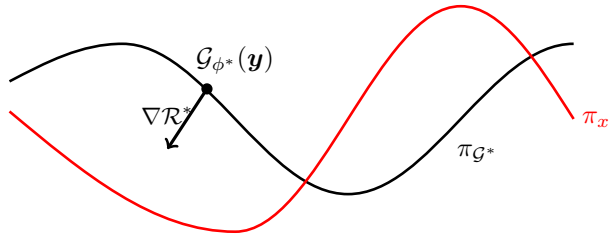
\begin{figure}[t]
\centering
\begin{tikzpicture}[xscale=1.50, yscale=1]
  \draw[line width=0.35mm] (0,0.5) sin (1,1) cos (2,0) sin (3,-1) cos (4,0) sin (5,1)
  plot[only marks] coordinates{(1,1)} (2.0,0.4) node [above = 0.7mm]{$ \G_{\phi^*}(\y)$} (4.15,-1.1) node[above = 4mm]{$\pi_{\G^*}$};
   \draw[->,very thick] (1.75,0.40) -- (1.4,-0.4) node[above = 2mm]{\small $\nabla \R^*$} ;
  \node at (1.75,0.40)  [circle,fill,inner sep=1.5pt]{};
  \draw[color=red, line width=0.35mm] (0,0.1) sin (2,-1.5) cos (3,0) sin (4,1.5) cos (5,0) (5.2,0) node {$\pi_x$};
 \end{tikzpicture}
 \caption{\footnotesize{A step of gradient descent applied to the initial point $\G_{\phi^*}(\y)$ is moving the point in the direction $\nabla \R^*(\G_{\phi^*}(\y))$ closer to the data distribution $\pi_x$}.}
 \label{fig:pushingdistr}
\end{figure}

We conclude this section by further discussing the content of Theorem \ref{thm:wassgradientflow}. As already noted, this theorem ensures that a gradient-descent step performed on $\R^*$ at $\x=\G_{\phi^*}(\y^{\delta})$ decreases the Wasserstein distance with respect to the ground-truth distribution $\pi_x$. Therefore, if the gradient-descent step to solve the variational problem \eqref{eq:inverseprob} is initialized with the reconstruction $\G_{\phi^*}(\y^{\delta})$, the next iterate gets pushed closer to the ground-truth distribution $\pi_x$. If we additionally use the same weak manifold assumption as in \cite{ar_nips} it is possible to prove that an optimal regularizer $\R^*$ is given by the distance function from the ground-truth manifold (see \cite{ar_nips}). In this case, if we additionally assume that the projection from $\G_{\phi^*}(\y^{\delta})$ to the manifold is unique, then the gradient of $\R^*$ in that point is a unit vector from $\G_{\phi^*}(\y^{\delta})$ to the unique projection point. Such consideration strengthens even more our claim that an iterate of gradient descent initialized in $\G_{\phi^*}(\y^{\delta})$ gets pushed closer to the ground-truth distribution $\pi_x$. A graphical representation of such effect is presented in Fig. \ref{fig:pushingdistr}.

\section{Additional numerical results}
Here, we provide a comparison of different algorithms on another test image from the Mayo-clinic low-dose CT challenge dataset \cite{mayo_ct_challenge} (See Figure \ref{ct_image_figure_mayo_supp} below). The purpose of this example is to demonstrate that the gain in performance achieved by UAR over the competing algorithms is consistent over different test images.  

\begin{figure*}[t]
\centering
		\subfigure[ground-truth]{
		\includegraphics[width=1.7in]{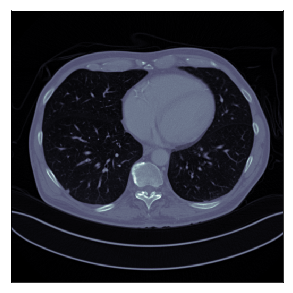}}
	\subfigure[FBP: 21.59, 0.24]{
		\includegraphics[height=1.7in]{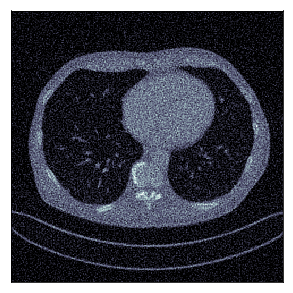}}
	\subfigure[TV: 29.16, 0.77]{
		\includegraphics[width=1.7in]{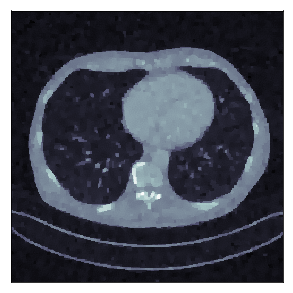}}\\
	\subfigure[U-net: 32.69, 0.87]{
		\includegraphics[width=1.7in]{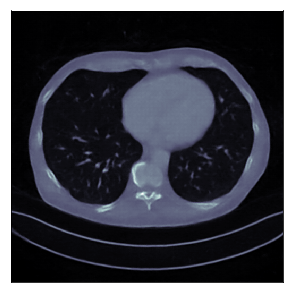}}
	\subfigure[LPD: 34.05, 0.89]{
\includegraphics[width=1.7in]{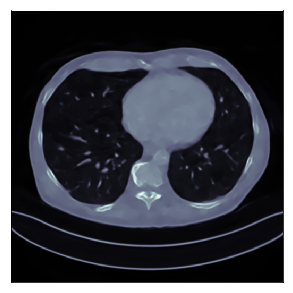}}
		\subfigure[ACR: 30.14, 0.83]{
		\includegraphics[width=1.7in]{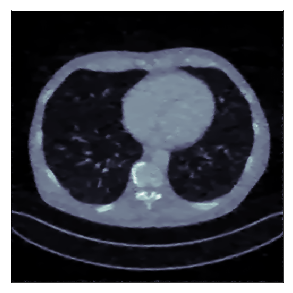}}\\
	\subfigure[AR: 32.14, 0.84]{
		\includegraphics[width=1.7in]{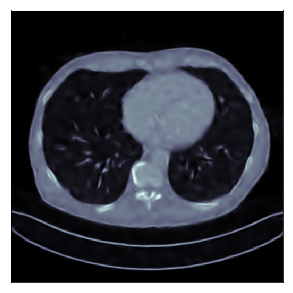}}
		\subfigure[UAR (end-to-end): 32.80, 0.86]{
		\includegraphics[width=1.7in]{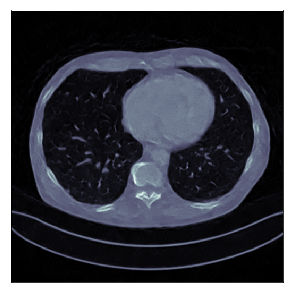}}
		\subfigure[UAR (refined): 33.15, 0.87]{
		\includegraphics[width=1.7in]{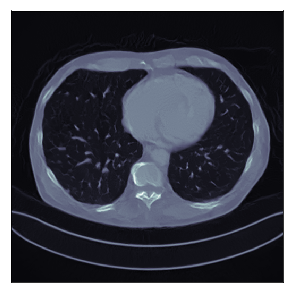}}
	\caption{Another numerical example on the Mayo clinic data \cite{mayo_ct_challenge}. As we see, UAR (refined) significantly outperforms AR and ACR, and achieves slightly better reconstruction quality than U-net-based post-processing, which is a supervised approach. To see the reduction in reconstruction time using UAR as compared to competing variational methods (such as TV, AR, and ACR), refer to Table \ref{sparse_ct_table} in Section \ref{numerics_sec}.}
	\label{ct_image_figure_mayo_supp}
\end{figure*}

\end{document}